\documentclass[10pt]{article}

\usepackage{thumbpdf,lmodern}

\usepackage{framed}

\usepackage{amsmath}
\usepackage{amssymb}
\usepackage{algorithmicx}
\usepackage{algorithm, algpseudocode}

\usepackage{graphicx}
\usepackage{pdflscape}
\graphicspath{{./figs/}}

\usepackage[authoryear,round]{natbib}
\usepackage{url}

\bibpunct{(}{)}{;}{a}{}{,}
\advance\textheight by \topskip
\let\proglang=\textsf
\newcommand{\pkg}[1]{{\fontseries{b}\selectfont #1}}

\usepackage{fancyhdr}
\begin{document}
\thispagestyle{empty}

{\centering

\vskip 0.3in
{\LARGE\textbf{bigMap: Big Data Mapping with parallelized t-SNE}}
\vskip 0.3in

Joan Garriga\textsuperscript{1}, Frederic Bartumeus\textsuperscript{1,2,3}
\vskip 0.1in

\textbf{1} Theoretical and Computational Ecology Laboratory,
\\ Centre d'Estudis Avan\c{c}ats de Blanes (CEAB-CSIC),
\\ Cala Sant Francesc, 14, 17300, Blanes, Spain
\\
\textbf{2} Centre de Recerca Ecol\`ogica i Aplicacions Forestals (CREAF),
\\ Cerdanyola del Vall\`es, 08193, Barcelona, Spain
\textbf{3}{Instituci\'o Catalana de Recerca, ICREA, Passeig Lluís Companys, 23, 08010, Barcelona, Spain }

\vskip 0.1in
E-mail: \url{jgarriga@ceab.csic.es}, \url{fbartu@ceab.csic.es}
\vskip 0.1in
URL: \url{http://theelab.net/}

\vskip 0.3in
\hrule


\begin{abstract}

We present an improved unsupervised clustering protocol particularly suited for large-scale structured data. The protocol follows three steps: a dimensionality reduction of the data, a density estimation over the low dimensional representation of the data, and a final segmentation of the density landscape. We improve all three steps by introducing: (i) a parallelized implementation of the well-known \textit{t-Stochastic Neighbouring Embedding} (t-SNE) algorithm that significantly alleviates some of its limitations, increasing its suitability for large data sets, (ii) a new \textit{adaptive Kernel Density Estimation} particularly coupled with the t-SNE framework to get accurate density estimates out of the embedded data, and (iii) a fast variant of the \textit{rainfalling watershed} algorithm to identify clusters within the density landscape. The whole mapping protocol is wrapped in the \textit{bigMap} \textit{R} package, together with visualization and analysis tools to ease the qualitative and quantitative assessment of the clustering.

\vskip 0.1in minus 0.05in
\textbf{Keywords}: big data, unsupervised clustering, t-SNE, parallel, \proglang{R}.

\end{abstract}

\vskip 0.1in
\hrule
\vskip 0.5in
}

\section*{Introduction}

A growing need in many fields of research is the development of tools to process and visualise large-scale structured data (LSSD). As an example, neurosciences and quantitative behaviour related fields, use large experimental data sets from model organisms (e.g. nematodes \citep{Nguyen:2016, Venkatachalam:2016}, fruit flies \citep{Berman:2014, Berman:2016}, zebrafish larvae \citep{Marques:2017}, mice, social insects \citep{Chandrasekaran:2011}), to profile and map behaviour at different levels of biological organization (i.e. genes, neurons, locomotion). One particular need in these studies is to devise unsupervised methods to infer the organizational principles and potential generative mechanisms underlying the data \citep{Gomez-Marin:2014} with minimal or no prior assumptions. Also, unsupervised visualization methods for the exploratory analysis of data are crucial in single-cell genomics and transcriptomics \citep{Kobak:2018} where improved experimental techniques generate gene expression data from tens to hundreds of thousands of cells (\textit{e.g.} \cite{Tasic:2018, Macosko:2015, Shekhar:2016}, 10XGenomics \url{http://10xgenomics.com}).

\textit{Mapping methods} (MM, \cite{Todd:2016}) constitute an effective approach to unsupervised clustering of LSSD. A MM starts with a non-trivial pre-processing step to convert raw data (usually unstructured data, \textit{e.g.} pictures, audio signals, video images) into a structured data set suitable for algorithmic analysis. Afterwards, a MM follows a multi-step clustering protocol over a low dimensional representation of the data. The particular techniques used at each step can vary. The dimensionality reduction of the data allows some downstream steps that otherwise would be computationally intractable. Additionally, embedding high dimensional data into a human-readable dimension (2D or 3D) simplifies the visualization and interpretation of the output clusters.

A successful result of MM was first reported for the study of adult \textit{Drosophila melanogaster} behaviour from video data \citep{Berman:2014, Berman:2016}. The MM included the use of the \textit{t-Distributed Stochastic Neighbouring Embedding} algorithm (t-SNE, \cite{Maaten:2008}) to reduce the high dimensional data to 2 dimensions, a fixed small bandwidth \textit{Kernel Density Estimation} (KDE, \cite{Terrell:1992}) to estimate a density function over the embedded space, and a \textit{watershed transform} (WT, \cite{Meyer:1994}) over the embedded space density landscape to get the final clustering.

\subsubsection*{t-SNE}

Dimensionality reduction techniques \citep{Lee:2007, Gisbrecht:2015} are mainly divided into linear embeddings \textit{e.g.} \textit{Principal Component Analysis} \citep{Hotelling:1993}), \textit{multidimensional scaling} \citep{Torgerson:1952}) focused on preserving the global structure of the data, and non-linear embeddings \textit{e.g.} \textit{Sammon mapping} \citep{Sammon:1969}, \textit{Isomap} \citep{Tenenbaum:2000}, \textit{Laplacian eigenmaps} \citep{Belkin:2001} focused on preserving the local structure in the data. In a context of unsupervised learning non-linear embedding looks more appealing because: (i) unveiling data structure at the local scale is fundamental as the input data is likely to be organized in a nonlinear manifold of much lower dimension; and (ii) when reduced to a human readable scale (\textit{i.e.} 2 o 3 first components) linear techniques might be droping off a significant ammount of information that might be crucial for visualization and analysis of data.

Among nonlinear dimensionality reduction techniques, t-SNE \citep{Maaten:2008, Maaten:2009a} shows up as an outstanding embedding algorithm for the visualization of high-dimensional data in a human-readable dimension space. The main driver of the embedding process is to preserve local pairwise similarities, \textit{i.e.} local similarities in the input space are mapped as close distances in the embedded space while moderate or large dissimilarities are not particularly preserved. The t-SNE achieves this by expressing the set of pairwise similarities into a joint probability distribution in both, the input (high dimensional) space and the embedded (low dimensional), and minimizing the divergence between the two distributions.

The major drawbacks of t-SNE (shared by many non-parametric dimensionality reduction methods) are: (i) the computational limits of the algorithm due to a quadratic time/space complexity, \textit{i.e.} beyond a few thousands of observations the embedding process becomes to slow to be of practical use, (ii) the uniqueness of the solution is far from being guaranteed, \textit{i.e.} solutions are highly dependent on the starting conditions, usually a random distribution generated from a particular seed value, and (iii) the qualitative/quantitative assessment of different solutions in an unsupervised context is not trivial at all. A minor (or not so) drawback is that t-SNE plots can sometimes be misleading if the idiosyncrasy of the algorithm is not well understood \citep{Wattenberg:2016}.

Herein, a lot of work has been devoted to improving the suitability of the t-SNE to LSSD, pushing efforts on two fronts: (i) development of platform-specific algorithmic implementations to make the most out of high-performance hardware (\textit{e.g.} multi-core tSNE \citep{Ulyanov:2016}, t-SNE-CUDA \citep{Chan:2018}, powerful implementations of the tSNE that parallelize some parts of the algorithm but do not resolve the iterative, dreadfully sequential, mapping process); and (ii) reexamination of existing algorithms under new, more effective, perspectives (\textit{e.g.} \cite{Maaten:2009b, Maaten:2014, Maaten:2017, Lee:2015, Linderman:2017, Linderman:2019, Arora:2018, Udell:2019}). Being both fronts of equal importance, our work aligns with the second one.

\subsubsection*{Our contribution}

We have reconsidered the t-SNE algorithm to make it more suitable to LSSD. The underlying assumption of our work is that LSSD usually convey a large amount of redundant evidence. Under this assumption, we approach the embedding problem with a \textit{divide and conquer} \citep{Cormen:1990} strategy, that is, breaking down the t-SNE into partial t-SNEs and setting the appropriate convergence conditions to combine the partial solutions into a global one. This approach results in a parallelized version of the t-SNE algorithm, namely the \textit{parallelized t-SNE} (ptSNE). The basic idea is to run several instances of the algorithm on different chunks of the data using an alternating scheme of short runs and mixing of the partial solutions. Based on the ptSNE, our goals are (i) to adapt and improve the t-SNE mapping protocol to LSSD, and (ii) to build a complete ready-to-use \textit{R} package for LSSD mapping.

\subsubsection*{data sets}

To show the performance of our mapping protocol we use the following data sets:

\begin{itemize}

\item MNIST (optical digits): A classical benchmark data set \citep{MNIST:2010, Lecun:1998} profusely used in image processing systems, supervised classification and dimensionality reduction algorithms. This is a large data set with $n=60000$ training images of handwritten digits. Images are encoded as integer vectors of 784 grey intensity levels (ranging in the interval 0-256) corresponding to an image resolution of 28x28 pixels (\textit{e.g.} Fig.~\ref{fig:MNIST}).

\item GMMx: A set of synthetically generated data sets sampled from a multi-dimensional \textit{Gaussian mixture model} (GMM). In particular we use GMM5 (5 dimensions, 32 Gaussian components, $n=200001$) and GMM7 (7 dimensions, 128 Gaussian components, $n=63998$), (\textit{e.g.} Fig.~\ref{fig:GMM7}).

\item dwt1005: A data set taken from the \textit{Sparse Matrix Collection} \citep{Davis:2009}, a data set repository for graph visualization. This data set represents a 3D mesh (Fig.~\ref{fig:dwt1005_ptSNE}, top-left) described as a fully connected undirected graph with 1005 nodes, where similarities are given as \textit{shortest path} distances between nodes.

\end{itemize}

\section*{\MakeLowercase{pt}SNE: parallelized \MakeLowercase{t}-SNE}

The t-SNE algorithm starts by transforming similarities (whatever measure of similarity) into a probability distribution \citep{Maaten:2008}. In the most common case, similarities are measured as pairwise euclidean distances among data points.

\paragraph{Similarities in the input (high dimensional) space, $\mathcal{X}\in \mathcal{R}^m$}

The similarity between observations $x_j$ and $x_i$, expressed as $\|x_i-x_j \| ^2$, is converted into the conditional probability $p_{j\mid i}$ given by a Gaussian kernel centered at $x_{i}$,

\begin{equation}
    p_{j\mid i} = \frac{\exp\left(-\beta_{i}\,\|x_i-x_j\| ^2\right)}{\sum_{k\neq i}\exp\left(-\beta_{i}\,\|x_i-x_k\|^2\right)}
\label{eq:h_cond_prob}
\end{equation}

\noindent with precision $\beta_{i}=1/\left(2\,\sigma_i^2\right)$. Decreasing values of $\beta_{i}$ induce a probability distribution of increasing entropy $H\left(p_{j|i}\right)$ and increasing \textit{perplexity}, defined as,

\begin{equation}
    Perp\left(p_{j|i}\right)=2^{H\left(p_{j|i}\right)}
\label{eq:perplexity}
\end{equation}

\noindent with a maximum value equal to $n-1$ (where $n$ is the data set size), corresponding to $\beta_{i}=0$ and a uniform distribution of similarities. Thus, as a preliminary step, t-SNE computes the values $\beta_{i}$ that result in a fixed perplexity for all $x_{i}$. The procedure to find $\beta_{i}$ is described in Supplemental File S1. Computing perplexity based similarities is a powerful transformation because it allows to control what is similarity in terms of spatial proximity without explicitly referring to any actual value of distance. In practical terms, low values of perplexity will unveil the local structure in the data, whereas high values of perplexity will enhance the emergence of the global structure at the cost of blurring the local structure. Thus, the perplexity sets a balance across the emergence of one or the other and must be tuned according to our requirements.

Afterwards, t-SNE computes a symmetric joint probability given by,

\begin{equation}
    p_{ij} = p_{ji} = \frac{p_{j\mid i}+p_{i\mid j}}{2n}
\label{eq:h_joint_prob}
\end{equation}

\noindent This ensures that $\sum_{i,j}\,p_{ij} = 1$ and $\sum_j\,p_{ij}>\frac{1}{2n}$ for all data points $x_i$, so that each data point plays its role in the embedding process \citep{Maaten:2008}.

\paragraph{Similarities in the output (low-dimensional) space, $\mathcal{Y}\in \mathcal{R}^d$, $d\in\{2, 3\}$}

The similarities between mapped data points $y_j$ and $y_i$, also expressed as $\|y_i - y_j\|^2$, are treated differently. A well-known issue of embedding processes is the so-called \textit{crowding problem} (\textit{i.e}. a surface at a given distance from a point in a high-dimensional space can enclose more points than those that can be accommodated in the corresponding low-dimensional area \citep{Maaten:2008}). This problem is alleviated using a heavy-tailed distribution to represent similarities in the low dimensional space, namely a Cauchy distribution (\textit{i.e.} a t-Student distribution with one degree of freedom). Therefore, the joint probabilities $q_{ij}$ are defined as,

\begin{equation}
q_{ij} = \frac{\left(1 + \| y_i-y_j \|^2\right)^{-1}}{\sum_{k\neq l}\left(1 + \| y_k-y_l \|^2\right)^{-1}}
\label{eq:l_joint_prob}
\end{equation}

The t-SNE uses a gradient descent method to find a low-dimensional representation of the data that minimizes the mismatch between $p_{ij}$ and $q_{ij}$. The cost function is defined as the Kullback-Leibler divergence between both distributions,

\begin{equation}
C = KL\left(P\|Q\right)=\sum_{i,j}p_{ij}\log\frac{p_{ij}}{q_{ij}}
\label{eq:cost_function}
\end{equation}

\noindent with a gradient with respect to the low-dimensional mapped positions given as \citep{Maaten:2008},

\begin{equation}
\frac{\delta C}{\delta y_i} = 4\sum_j\left(p_{ij}-q_{ij}\right)\left(y_i-y_j\right)\left(1+\|y_i-y_j\|^2\right)^{-1}
\label{eq:cost_gradient}
\end{equation}

\begin{figure}[!t]\centering
\includegraphics[width=11.0cm, height=7.0cm]{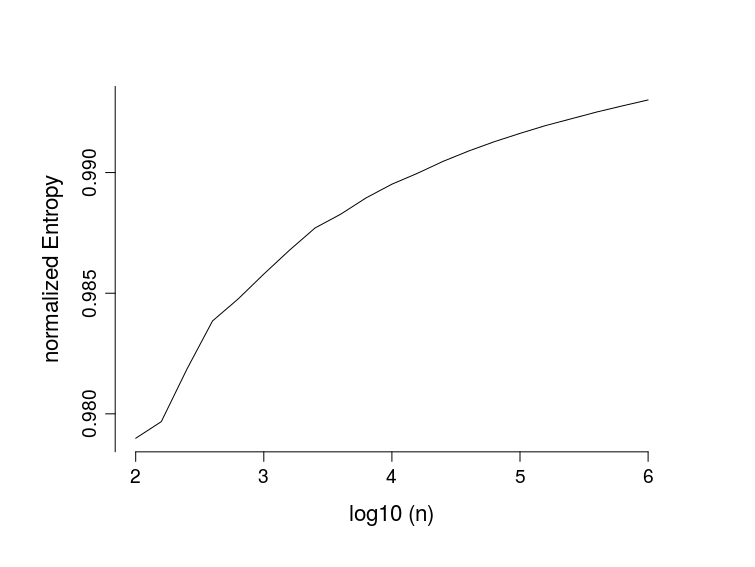}
\caption{\textbf{The big crowding problem}. Normalized entropy of the distribution of similarities (as a measure of the average homogeneity on pairwise data point similarities) for data sets of increasing size sampled from the same bivariate (2D) Gaussian kernel. Entropy values are average values from 100 samples.}
\label{fig:BCproblem}
\end{figure}

\subsection*{The big crowding problem}

t-SNE holds an implicit dependence on the size $n$ of the data set. The reason is that the t-SNE transforms similarities into a joint probability distribution with a finite amount of probability mass to be allocated among all pairwise distances, which grow with $n\,\left(n-1\right)$. Therefore, as $n$ grows, the values of similarity must be lower on average and tend to be more homogeneous. We can show this by considering a subset of data points sampled from a bivariate (2D) kernel centred at a data point $i$ and computing the similarities $p_{j|i}$ (Eq. \ref{eq:h_cond_prob}). Afterwards, we can compute the normalized entropy of the resulting distribution. The normalized entropy is a measure of the average homogeneity such that the closer to 1 the entropy the more homogeneously distributed are the similarities.  Repeating this process with subsets of increasing size, we observe how the entropy tends to 1 as the size grows (Fig.~\ref{fig:BCproblem}). For large data sets, this fact generates undesired effects that we discuss throughout this work and that we generically call the \textit{big crowding problem}.

\subsection*{Pseudo-normalized cost function}

A first effect of the \textit{big crowding problem} is that the cost function (Eq.~\ref{eq:cost_function}) holds itself an implicit dependence on $n$. Let's consider the average similarity of a random embedding of $n\left(n-1\right)$ pairwise distances. Based on a Cauchy distribution, the average similarity is,

\begin{flalign}
\nonumber
\langle q_{ij}\rangle
&= \frac{\left(1+\langle\|y_i-y_j\|\rangle^2\right)^{-1}}{\sum_{j\neq i}\left(1+\|y_i-y_j\|^2\right)^{-1}} \\
\nonumber
&= \frac{\left(1+\langle\|y_i-y_j\|\rangle^2\right)^{-1}}{n\,\left(n-1\right)\left(1+\langle\|y_i-y_j\|\rangle^2\right)^{-1}} \\
&= \frac{1}{n\,\left(n-1\right)}
\label{eq:average_q}
\end{flalign}

\noindent Plugging the average similarity (Eq.~\ref{eq:average_q}) into Eq.~\ref{eq:cost_function} we have an average cost,

\begin{flalign}
\nonumber
\langle C \rangle
&\propto -\log \langle q_{ij} \rangle \sum_{i,j} p_{ij} \\
&\propto \;\log n + \log\left(n-1\right)
\label{eq:average_cost}
\end{flalign}

\noindent where we have also dropped the term $\sum_{i,j}p_{ij}\log p_{ij}$, which is constant along the optimization of the embedding.

It turns out that the expression in Eq.~\ref{eq:average_cost} is the cost of a uniform distribution of similarities, \textit{i.e.} the cost of a uniform embedding of $n\,\left(n-1\right)$ pairwise distances, expressing that all data points are equally similar. While it is not feasible to arrange $n\,\left(n-1\right)$ uniform pairwise distances in 2D, such a uniform distribution constitutes the worst possible embedding with respect to $P$, be $P$ what it may. Thus, Eq.~\ref{eq:average_cost} is an upper bound in terms of $KL\left(P\|Q\right)$ divergence and we can define a pseudo-normalized cost function as,

\begin{equation}
C = -\frac{\sum_{i,j}p_{ij}\log q_{ij}}{\log n+\log\left(n-1\right)}
 = -\frac{H\left(P, Q\right)}{H\left(P, U\right)}
\label{eq:normalized_cost}
\end{equation}

\noindent In terms of information theory this is the \textit{normalized cross-entropy} of distributions $P$ and $Q$, that is, the average cost of coding $P$ as $Q$ relative to the worst-case cost, which is the cost of a uniform embedding of $n\,\left(n-1\right)$ pairwise distances.

This pseudo-normalized cost yields always a value very close to 1 for a random initial mapping, and allows a fair comparison of results from different runs (as long as perplexity is fixed).

\subsection*{Parallelized implementation}

The ptSNE algorithm runs several instances (independent threads) of the t-SNE on different chunks of data (partial t-SNEs) using an alternating scheme of short runs and mixing of the partial solutions (Fig. \ref{fig:ptSNE_scheme1}). The algorithm starts by randomly allocating the data points in the low-dimensional space (a disk area of radius 1). The iteration cycle is arranged into \textit{epochs} with a short number of iterations each. After each epoch, the solutions of the partial t-SNEs are pooled together, resulting in a new global embedding. A new epoch is started by sending to each thread a new chunk of data the starting positions of which result from a mixture of the previous independent solutions. The run and mixing scheme is governed by the following parameters (Supplemental File S2):

\begin{figure}[!t]\centering
\includegraphics[width=13.5cm, height=6.5cm]{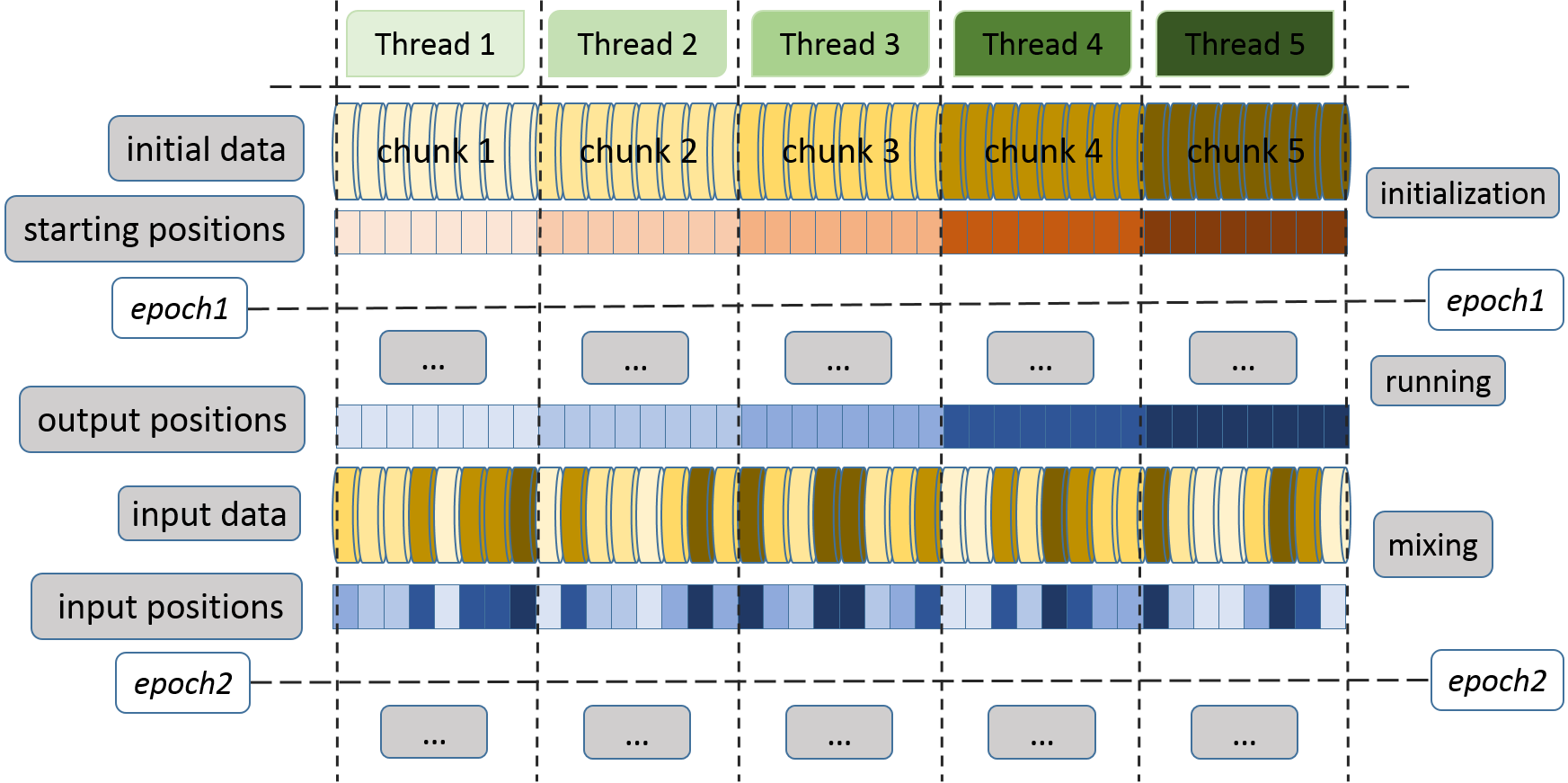}
\caption{\textbf{ptSNE basic parallelization scheme}. The parallelized implementation runs a number (5 in this example) of instances of the t-SNE algorithm in an alternating scheme of short runs and mixing of partial solutions. Each run-and-mix phase is an epoch. In this example, each thread iterates on a single chunk of data, starting with random mapping positions. After a number of iterations the partial t-SNEs are pooled together and mixed. A new epoch is started with each thread iterating on a new chunk of data and starting with current mapping positions.}
\label{fig:ptSNE_scheme1}
\end{figure}

\begin{figure}[!th]\centering
\includegraphics[width=12cm, height=5.5cm]{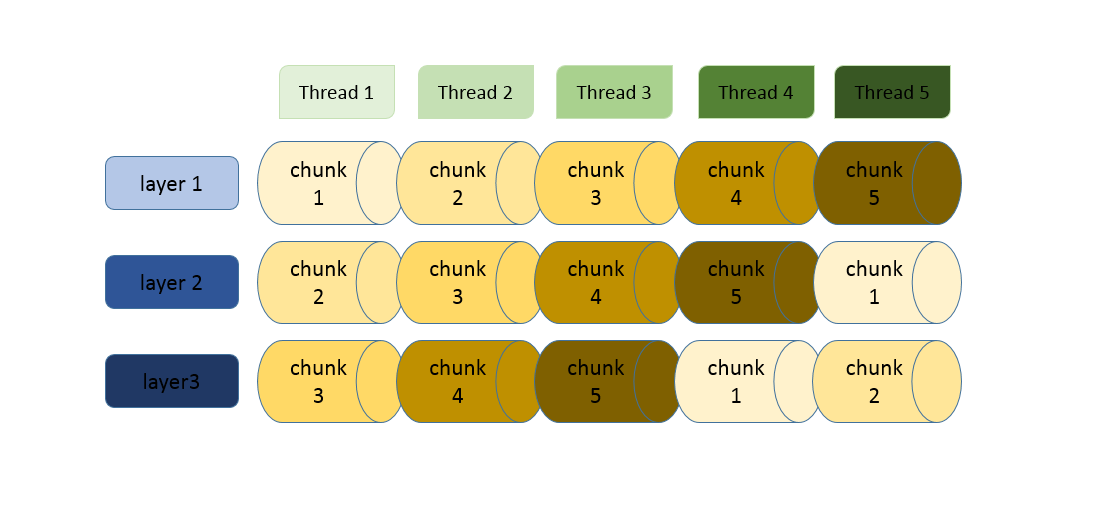}
\caption{\textbf{ptSNE parallelization scheme with 3 layers}. Each thread iterates on 3 chunks of data sharing each one of them with successive threads. Common data points create a link between the partial solutions that favours convergence. As each data point is running on 3 different threads we get 3 different mapped positions for each one. After pooling all partial solutions we get 3 global mappings (layers).}
\label{fig:ptSNE_scheme3}
\end{figure}

\subsubsection*{threads}

The number of threads is the number of partial t-SNEs that will run. The ptSNE splits the data set into this number of elementary chunks, so that, the larger the number of threads, the faster the computation of the final solution. Note that the number of threads must not necessarily match the number of physical cores available. Indeed, it can be higher (what is known as multi-threading). Using multi-threading to run ptSNE on multiprocessor systems can yield significant reductions in computational time.

\subsubsection*{layers}

In the most simple scheme (Fig. \ref{fig:ptSNE_scheme1}), each thread runs a single chunk of data and the partial solutions are pooled together, mixed and chunked again to start a new epoch. However, the key for convergence of the partial t-SNEs towards a common solution is to let the threads share some data. Thus, instead of running a single chunk of data, each thread runs as many chunks of data as specified by \textit{layers}, and all the threads are chained cyclically sharing a chunk of data with the $layers-1$ subsequent threads in the chain. For instance, with $threads=5$ and $layers=3$ (Fig. \ref{fig:ptSNE_scheme3}) the ptSNE would pool chunks 1, 2, and 3 into thread 1, chunks 2, 3 and 4 into thread 2, and so on, up to chunks 5, 1 and 2 into thread 5). This chained link between all threads is what favours convergence. Another consequence of the overlapping between threads is that each data-point is taking part in multiple (3 in the former example) partial t-SNEs. Thus, after pooling all partial solutions, we have as many global solutions as \textit{layers} (3 in our example).

\subsubsection*{thread-size}

The relation $layers/threads$ determines the thread-size $z=n\,*\,layers/threads$ (where $n$ is the data set size). As t-SNE is of order quadratic to the size of the data set, by making $z \ll n$ we overcome the unsuitability of the t-SNE algorithm for large-scale data sets. The $layers/threads$ ratio represents a trade-off between computational time and optimality of the global solution. The closer is the ratio to 1, the larger the percentage of data points used in each partial t-SNE and the more robust and comprehensive is the solution. This comes at the cost of a much larger computational cost. However, for large data sets ($n>10^4$), ptSNE yields a good global solution even with values of $z$ as low as 1\% of $n$.

\subsubsection*{epochs}

The run and mixing scheme is cyclical and each cycle is structured in three phases involving a master and several worker processes (the threads): (i) the master mixes the data and defines the data chunks; (ii) each worker runs a partial t-SNE with the chunk of data that it has been assigned; (ii) the master pools the partial solutions from the workers into a global solution. This sequence constitutes one \textit{epoch}, and the ptSNE involves a predetermined number of epochs. The running phase is usually kept short (default setup) to avoid too divergent partial solutions at the end of the epochs (particularly in the initial stages of the algorithm).

\subsubsection*{rounds}

The number of epochs is set to $\sqrt{n}$, and the number of iterations per epoch (epoch length) is set to $\sqrt{z}$. Scaling the epoch length to the thread-size avoids getting too divergent solutions from each partial t-SNE. This setup, with $\sqrt{n}$ epochs and $\sqrt{z}$ iterations per epoch, is a \textit{round}. In general, the algorithm reaches a stable solution in one single round. If not, the ptSNE can run extra rounds to refine the mapping, although the improvement achieved is usually low for the computational time required. The decision of running extra rounds can be easily assessed by evaluating the embedding cost and embedding size functions as they both show a flat line when the solution is stable, \textit{e.g.} Fig.~\ref{fig:rounds}.

\subsection*{Internal parametric configuration}

Original implementations of the t-SNE algorithm (\textit{e.g.} R \cite{Krijthe:2015}, python \cite{sklearn:2011} and Matlab \cite{Maaten:2009a}) start with a random embedding, by sampling mapped positions from an isotropic Gaussian, and update them iteratively using the following expression,

\begin{equation}
 \mathcal{Y}^{\left(t\right)} =\mathcal{Y}^{\left(t-1\right)} +\eta^{\left(t \right)}\frac{\delta C}{\delta \mathcal{Y}} +\alpha^{\left(t\right)}\left(\mathcal{Y}^{\left(t-1\right)}-\mathcal{Y}^{\left(t-2\right)}\right)
\label{eq:mapping_update}
\end{equation}

\noindent where $\mathcal{Y}^{\left(t\right)}$ is the set of mapped positions at iteration $t$, $\eta^{\left(t \right)}$ is a learning rate factor, and $\alpha^{\left(t \right)}$ (an external parameter) is a relatively large momentum factor, both factors using an adaptive scheme to speed up the optimization. A further strategy to enhance the optimization is the so-called \textit{early exaggeration} \citep{Maaten:2008}, a multiplying factor of the distances. Altogether, the t-SNE presents a contrived internal parametric setup specifically designed to afford a fast generation of tight apart initial clusters and fast convergence. Conversely, a parallelized implementation based on chunks of data demands a smooth clustering evolution, particularly avoiding too early arrangements of the clusters as this would compromise the convergence among partial solutions. For this reason, we use a much simpler parametric arrangement where the learning-rate $\eta^{\left(t \right)}$ is the one-and-only internally fixed parameter. However, for the learning-rate, we use an auto-adaptive scheme controlling for two implicit biases present in the update of the mapping positions $\Delta\mathcal{Y}=\mathcal{Y}^{t}-\mathcal{Y}^{\left(t-1\right)}$:

\begin{enumerate}

\item \textbf{The size of the embedding area}: As the the mapping positions are updated the size of the embedding area $\mathcal{E}$ must grow to accommodate moderate and large dissimilarities. However, as the size of the embedding grows the updates themselves $\Delta\mathcal{Y}$ decrease (\textit{i.e.} $\lim_{\mathcal{E}\rightarrow\infty}\Delta\mathcal{Y}=0$). This is due to the factor $\left(y_i-y_j\right)\left(1+\|y_i-y_j\|^2\right)^{-1}$ in the cost gradient (Eq.~\ref{eq:cost_gradient}) with,

\begin{displaymath}
\lim_{\|y_i-y_j\|\rightarrow\infty}\frac{\|y_i-y_j\|}{\left(1+\|y_i-y_j\|^2\right)} = 0
\end{displaymath}

\noindent independently of the matching between $P$ and $Q$. Thus, we compensate this effect by making,

\begin{displaymath}
\eta^{\left(t\right)} \propto \left(\mathcal{Y}^{\left(t\right)}_{max}-\mathcal{Y}^{\left(t\right)}_{min}\right)
\end{displaymath}

\item \textbf{The size of the data set}: The larger is the size of the data set the smaller are, on average, the position updates $\Delta\mathcal{Y}$. This is an effect of the \textit{big crowding problem}, so that a larger amount of data points leads to smaller values, on average, of the factor $\left(p_{ij}-q_{ij}\right)$ in the cost gradient (Eq.~\ref{eq:cost_gradient}). We can better understand this effect by combining Eq.~\ref{eq:mapping_update} and~\ref{eq:cost_gradient} where we see that,

\begin{displaymath}
\frac{\delta}{\delta n}\,\Delta y_{i}
\,\propto\,\frac{\delta}{\delta n}\,\frac{\delta\,C}{\delta y_{i}}
\,\propto\,\frac{\delta}{\delta n}\,\sum_{j\neq i}\left(p_{ij}-q_{ij}\right)
\end{displaymath}

\noindent that is, for a given mapped position $y_i$, this effect runs through the $\left(n-1\right)$ data points $y_{j\neq i}$. Then, recalling Eq.~\ref{eq:average_cost}~and~\ref{eq:normalized_cost}, it seems reasonable to compensate this effect on the learning-rate by making,

\begin{displaymath}
\eta \propto \log(n-1)
\end{displaymath}

\end{enumerate}

Combining the above two considerations, we get an expression for the learning-rate that results in a smooth and appropriate auto-adaptive scheme,

\begin{equation}
\eta^{\left(t\right)} = \log\left(n-1\right)\; \frac{\mathcal{Y}^{\left(t\right)}_{max}-\mathcal{Y}^{\left(t\right)}_{min}}{2}\,
\label{eq:learning_rate}
\end{equation}

On the initial stages of the optimization, the factor $\left(p_{ij}-q_{ij}\right)$ will likely be large while the difference $\mathcal{Y}^{\left(t\right)}_{max}-\mathcal{Y}^{\left(t\right)}_{min}$ is small, thus the learning-rate will mitigate the impact of the strong attraction/repulsion forces originated during this critical moment. Along the curse of the optimization, the difference $\mathcal{Y}^{\left(t\right)}_{max}-\mathcal{Y}^{\left(t\right)}_{min}$ will get larger increasing the learning rate to compensate the decreasing attraction/repulsion forces.

\section*{Clustering}

Downstream steps in our mapping protocol consist on (i) estimating a density function over the low dimensional embedding of the data set (the output of the ptSNE) and (ii) finding an optimal partition of the estimated density landscape.

\subsection*{Perplexity-adaptive Kernel Density Estimation}

The \textit{Kernel Density Estimation} (KDE) is a family of non-parametric methods to estimate univariate or multivariate probability density functions based on finite data samples. In the most common form, a KDE is a mixture of kernels (a non-negative function that has zero mean and integrates to one, \textit{e.g.} a Gaussian kernel) with a smoothing parameter called the bandwidth (\textit{e.g.} the standard deviation of a Gaussian kernel). A general approach to improve kernel density estimates is to use adaptive bandwidth estimators. Methods based on this approach are known as \textit{adaptive kernel density estimators} (AKDE) and one such AKDE is the \textit{sample smoothing estimator} \citep{Terrell:1992},

\begin{equation}
 \hat{f}\left(c\right) = \frac{S}{n}\,\sum_{j=1}^{n}\frac{1}{h\left(y_{j}\right)^d}\;K\left(\frac{c-y_{j}}{h\left(y_{j}\right)}\right)
\label{eq:sampleSmoothingEstimator}
\end{equation}

The \textit{sample smoothing estimator} is a mixture of identical but locally scaled kernels centred at each observation, where the kernel bandwidth $h\left(\cdotp\right)$ is dependent on the sample mapped point $y_j$. The estimated density is a rasterized function on a regular grid over the embedding area with cells $c$ of size $S$. The advantage of this estimator is that if $K\left(\cdotp\right)$ is a density then so is $\hat{f}\left(c\right)$.

In general, KDE approximations put the focus on the underlying density function to estimate an optimal bandwidth, based on the asymptotic mean squared error \citep{Terrell:1992}. Our proposal to scale the kernels is to put the focus on similarities, and estimate local bandwidths based on a given value of perplexity (\textit{i.e.} we just borrow the perplexity based approach used for the t-SNE algorithm in the high dimensional space and apply it to convert distances into similarities in the low dimensional space). This constitutes a coherent strategy that makes the similarity-based approach a backbone of the MM. Thus, we base our \textit{sample smoothing estimator} on locally scaled bivariate Gaussian kernels defined as,

\begin{equation}
K_{j}\equiv \;K\left(\frac{c-y_{j}}{h\left(y_{j}\right)}\right)
\equiv \frac{1}{\pi}\,\exp\left(-\beta_{j}\,\|c-y_{j}\|^2\right),\quad\forall y_{j}\in\mathcal{Y}
\label{eq:adaptiveGaussianKernel}
\end{equation}

\noindent with bandwidth,

\begin{equation}
h\left(y_{j}\right)^2 \equiv 2\,\sigma_{j}^2 = \frac{1}{\beta_{j}}
\label{eq:adaptiveBandwidth}
\end{equation}

\noindent where $\mathcal{Y}$ are the mapped data points, and the precisions $\beta_{j}$ are found through the same procedure used in the high dimensional space (Supplemental File S1). Herein, we refer to the density estimation algorithm based on this estimator as the \textit{perplexity-adaptive kernel density estimation} (paKDE).

Local kernels with perplexity-based bandwidths adapt themselves to the neighbouring density around each mapped data point. Note that given the value of perplexity, $\beta_{j}$ will be such that the more dense the region where $y_{j}$ is mapped, the lower the bandwidth $h\left(y_{j}\right)$. Therefore, $K_{j}\left(\cdotp\right)$ will concentrate more density in the neighbourhood of $y_{j}$. As a result, the density function will become strongly constrained to the shape of the embedding in areas of high density but looser in areas of low density.

\subsection*{The Water-track Transform}

Density landscape partitioning involves evaluating peaks and valleys within the landscape to find connected areas across different scales. A particular framework for landscape partitioning is the \textit{watershed transform} (WT) \citet{Meyer:1994}. This framework is mainly devised for image segmentation, \textit{i.e.} the process of partitioning a digital image in meaningful or homogeneous segments by first converting images into a topographic relief based on pixel intensity, or intensity gradients. To segment the topographic landscape in distinct regions or areas, WT algorithms use different techniques that are mainly divided into \textit{flooding} algorithms \citep{Vincent:1991} and \textit{rainfalling} (or \textit{steepest descent}) algorithms \citep{Stoev:2000, DeBock:2005}.

\begin{figure}[tp]\centering
\includegraphics[width=14.5cm, height=3.0cm]{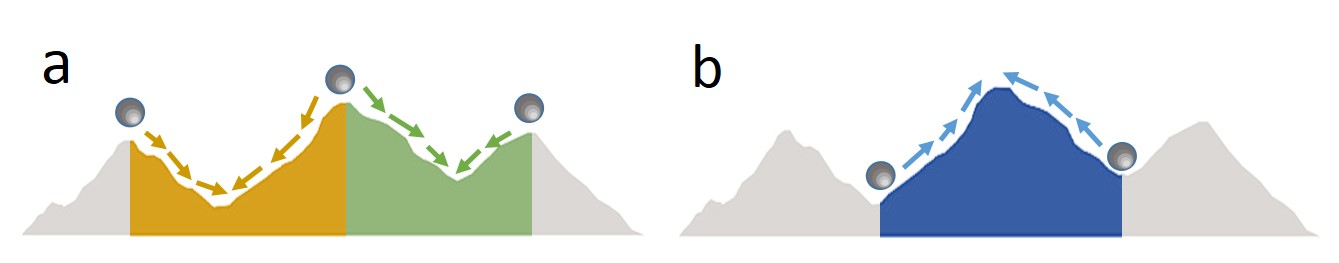}
\caption{\textbf{WTT algorithm.} Depiction of the classic rainfalling variants of the WT (a) versus the WTT algorithm principle (b).}
\label{fig:WTT}
\end{figure}

The general principle in watershed algorithms is to identify segments as the valleys in the landscape and the surrounding mountain rims as the boundaries of the segments (Fig.~\ref{fig:WTT}a). This is not what we should be looking for when dealing with similarity landscapes, where peaks are representing maximum similarity among data points. Because of this, we developed an algorithm that looks for peaks (local maxima) and identifies the river beds (water tracks) in the surrounding valleys as the boundaries of the segments (Fig.~\ref{fig:WTT}b). Hence, each local maximum in the density landscape identifies a clustered region that embraces all the area such that, climbing up the gradient of probability density, leads to itself (Fig.~\ref{fig:WTT}b). The result is similar to the output of a gradient-ascent algorithm (\textit{e.g.} mean-shift \citet{Fukunaga:1975}) though the implementation is quite different. We name this inverted variant of the rain-falling algorithm the \textit{water-track transform} (WTT) as, to the most of our knowledge, there is not a similar implementation described in the literature.

The input of the WTT is the density function, a raster function on the paKDE grid. Consequently, the output (the clustering labels) is given at grid-cell level. The implementation of the WTT is very simple: sort cells by density in descending order and sequentially label them as the highest neighbouring cell that has already been labelled; if no neighbouring cell has already been labelled (as it is the case for the first one), the cell must be a peak, thus it is labelled as a new cluster. In other words, think of coloured water flowing in all directions from the top of each peak and colouring all downstream cells with the colour assigned to that peak. To set the neighborhood of the cells we consider an 8-connectivity scheme \citep{Meyer:1994}.

A minor shortcoming of WT algorithms is the existence of plateaus (regions where the gradient is not defined). As cells with equal height end up sorted unpredictably, the algorithm will likely reach a plateau by hitting first a cell that is not at the border of the plateau. In this case, all its neighbours are still unlabelled and that cell will be wrongly labelled as a new cluster. Thus, when hitting a cell with unlabelled neighbourhood, we additionally check if the cell is indeed the highest of its neighbourhood. If it is not, it must belong to a plateau, and it is moved down the sorted list to the last position of the cells with that height. This step is repeated for all the cells with equal height until the algorithm hits a cell in the border of the plateau that can be correctly labelled. This process ends up labelling all the cells of the plateau. As clusters are labelled by decreasing density, the labels convey a relative significance of the clusters. Also, note that clustering labels are given at grid-cell level. More details are given in Supplemental File S3.

Given the value of perplexity in ptSNE and paKDE, the output of the WTT is the clustering labels at the finest grain level. To achieve coarser classifications, we can either repeat the whole MM with lower values of perplexity or apply \textit{ad-hoc} merging techniques.

The computational cost of this algorithm depends on the grid resolution $g$ (usually $g\ll n$), and it is approximately of $\boldsymbol{O}\left(g\,\log g\right)$, as it is basically a sorting algorithm plus some extra computation to solve plateaus.

\section*{Results}

We analyze the performance of the ptSNE with respect to the main two parameters of the algorithm: (i) the \textit{perplexity}, a neighbouring parameter accounting for for the degree of similarity among data points, and (ii) the \textit{thread-size} (determined by the \textit{layers-to-threads} ratio), accounting for the amount of information contained and overlapped in each partial t-SNE. In summary, the combination of perplexity and thread-size sets the balance between capturing local and global structure.

\begin{figure}[!p]\centering
\begin{tabular}{ccc}
\includegraphics[width=3cm, height=3cm]{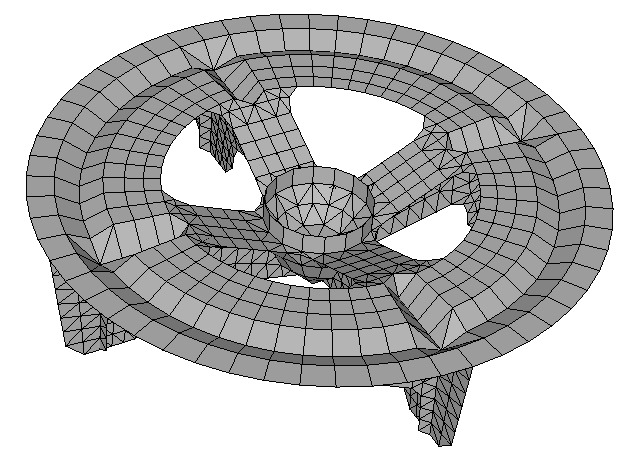}
&
\includegraphics[width=5cm, height=3cm]{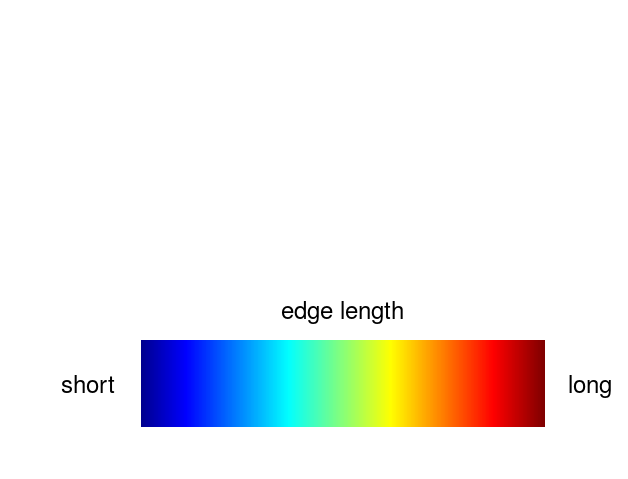}
&
\\
\includegraphics[width=4.8cm, height=3.5cm]{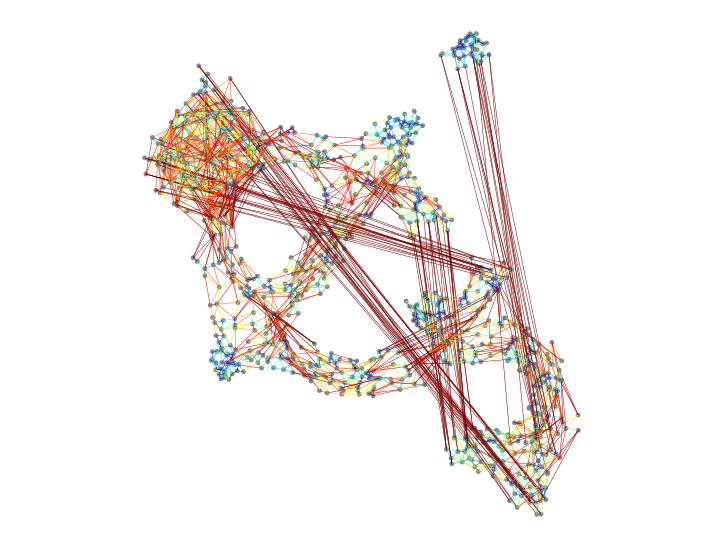}
&
\includegraphics[width=4.8cm, height=3.5cm]{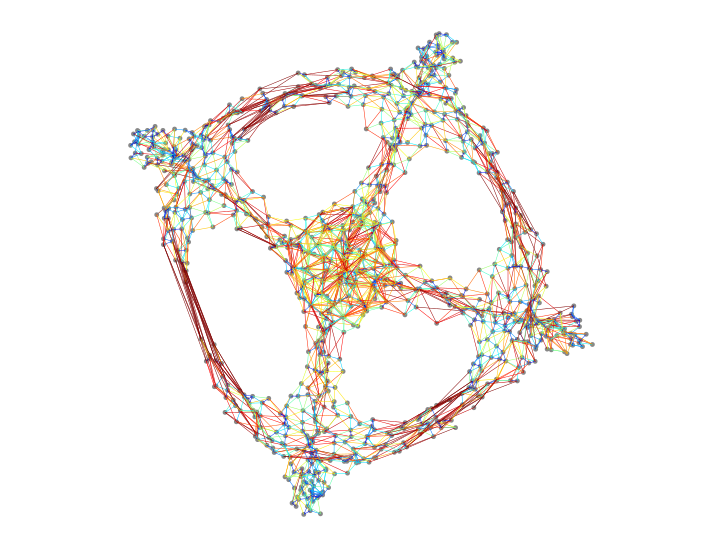}
&
\includegraphics[width=4.8cm, height=3.5cm]{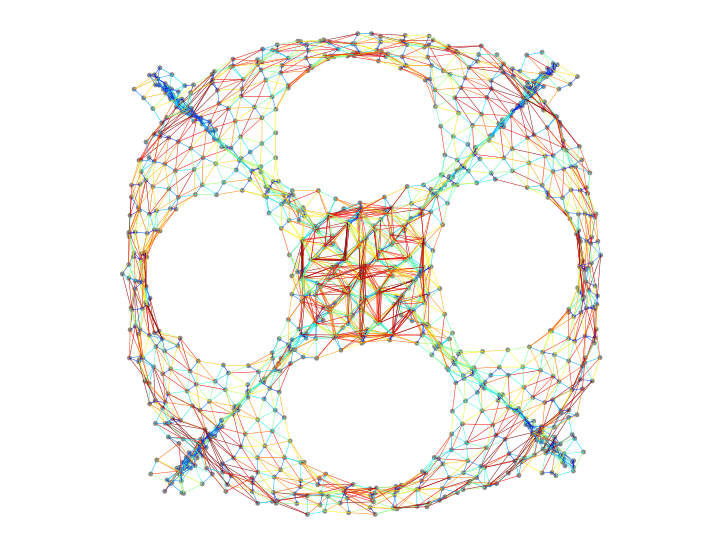}
\\
\includegraphics[width=4.8cm, height=3.5cm]{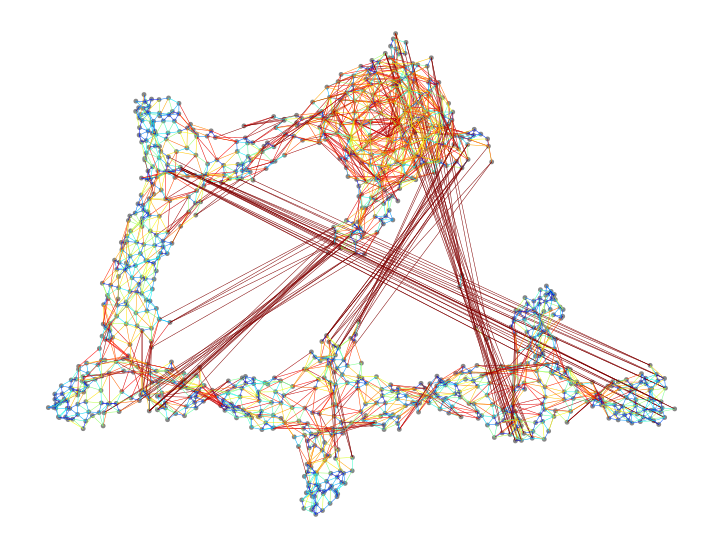}
&
\includegraphics[width=4.8cm, height=3.5cm]{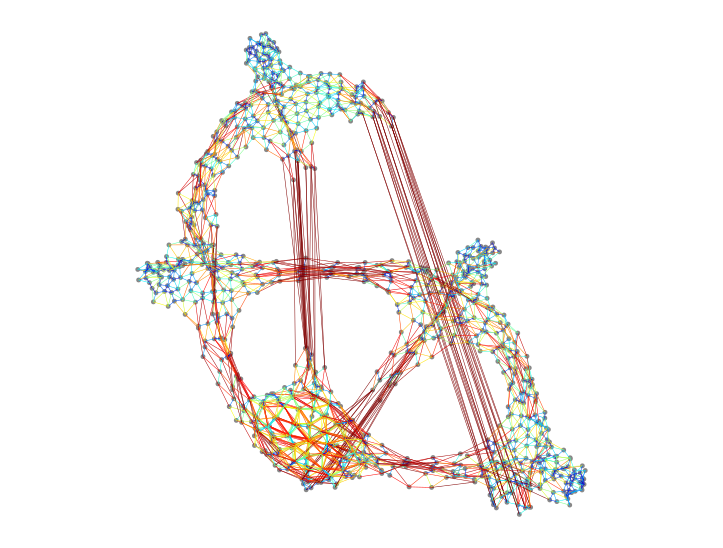}
&
\includegraphics[width=4.8cm, height=3.5cm]{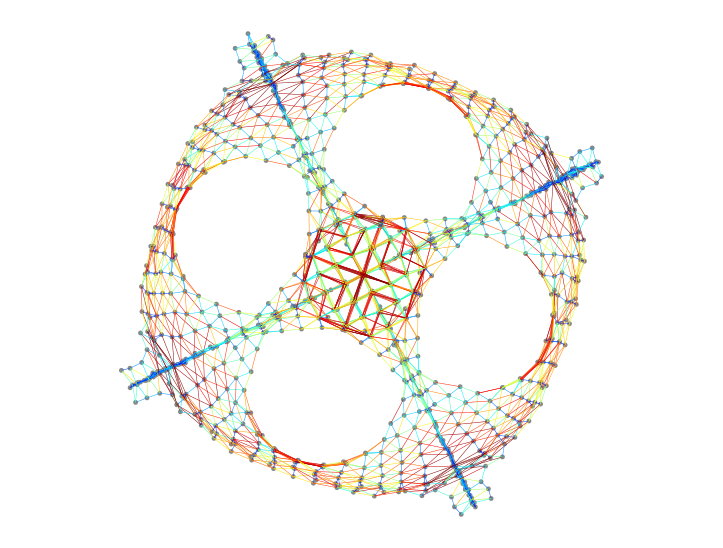}
\\
\includegraphics[width=4.8cm, height=3.5cm]{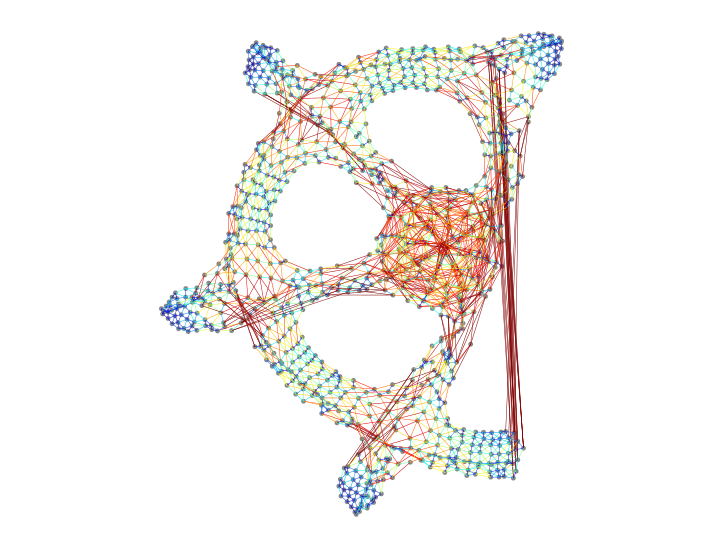}
&
\includegraphics[width=4.8cm, height=3.5cm]{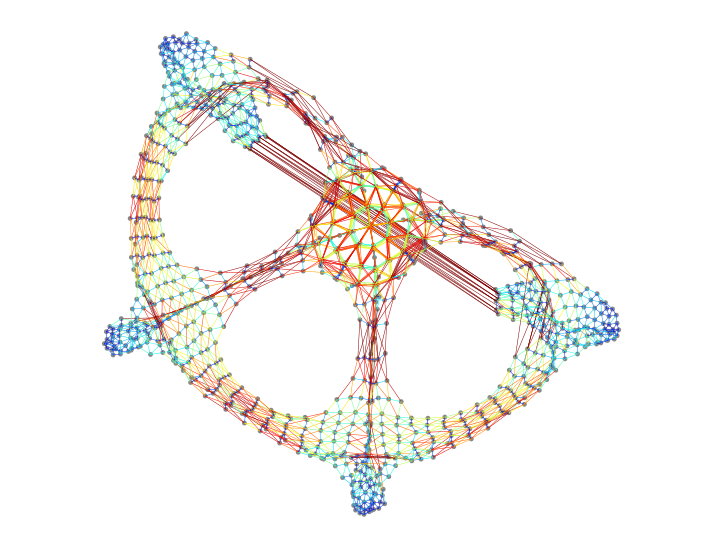}
&
\includegraphics[width=4.8cm, height=3.5cm]{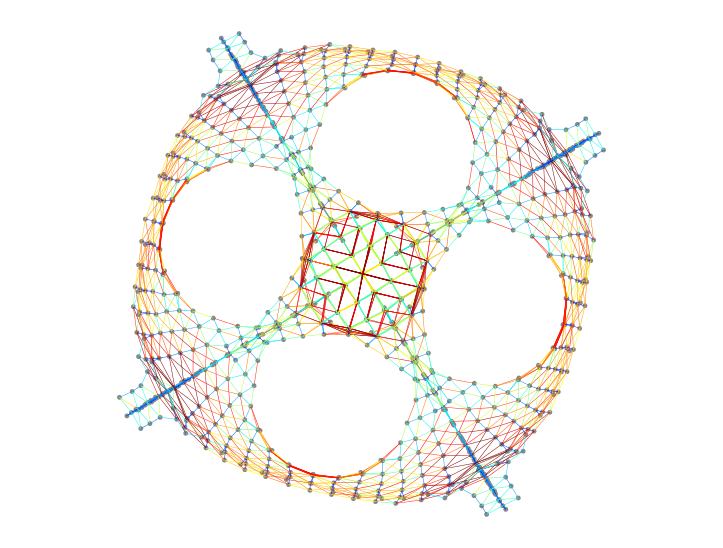}
\end{tabular}
\caption{\textbf{Effect of perplexity, threads and layers} The \textit{dwt1005} data set represents a 3D mesh (top-left) described as a fully connected undirected graph, with distances given as shortest path distances ($n=1005$). Bottom grid: ptSNE output performed with 10 threads for different number of layers ($\{ 2, 4, 8 \}$, \textit{i.e} a thread-size equivalent to 20, 40 and 80\% of $n$, top to bottom) and different perplexities ($\{ 50, 200, 800 \}$, left to right). All runs performed with 4 rounds. Colours encode relative lengths of the mapped edges (top-right).}
\label{fig:dwt1005_ptSNE}
\end{figure}

Because of the unsupervised nature of the algorithm, we rely on no other means than visually inspecting the output of the ptSNE. For this reason, we use the \textit{dwt1005} data set, a graph with known structure. Dimensionality reduction and, in particular, the t-SNE algorithm, has been proofed successful as a method for the visualization of graphs \citep{Kruiger:2017}. The analysis of the output of the \textit{dwt1005} data set under different parametric configurations highlights many interesting insights about the performance of the ptSNE (Fig.~\ref{fig:dwt1005_ptSNE}):

\begin{itemize}

\item Increasing the perplexity (Fig.~\ref{fig:dwt1005_ptSNE}, left to right, $ppx=\{50, 200, 800\}$) leads to a better definition of the global structure. Low values of perplexity favour the emergence of local structure but the algorithm is likely to reach suboptimal solutions. The reason is that at the initial stages of the optimization, the embedding will likely show strong foldings of the structure that can not be resolved unless long distant nodes are also playing its role in the mapping, and this is only possible with high values of perplexity. Therefore, a good strategy is to perform multiple runs of the ptSNE, starting with relatively high values of perplexity and lowering them gradually as long as the global structure is seamlessly preserved and local structuration starts to emerge.

\item Increasing the thread-size (Fig.~\ref{fig:dwt1005_ptSNE}, top to bottom, $layers/threads=\{ 0.2, 0.4, 0.8 \}$), leads to higher precision embedding, that is, the structure, either local or global, is sketched with higher precision. For instance, the results for perplexity 800 (right column in Fig.~\ref{fig:dwt1005_ptSNE}) clearly show that lowering the thread-size (bottom to top) implies a loss of information that leads to a worst mapping.

\item High values of both parameters will favour the robustness of the output through different runs (although possibly rotated).

\end{itemize}

\begin{figure}[!th]\centering
\includegraphics[width=15cm, height=5.3cm]{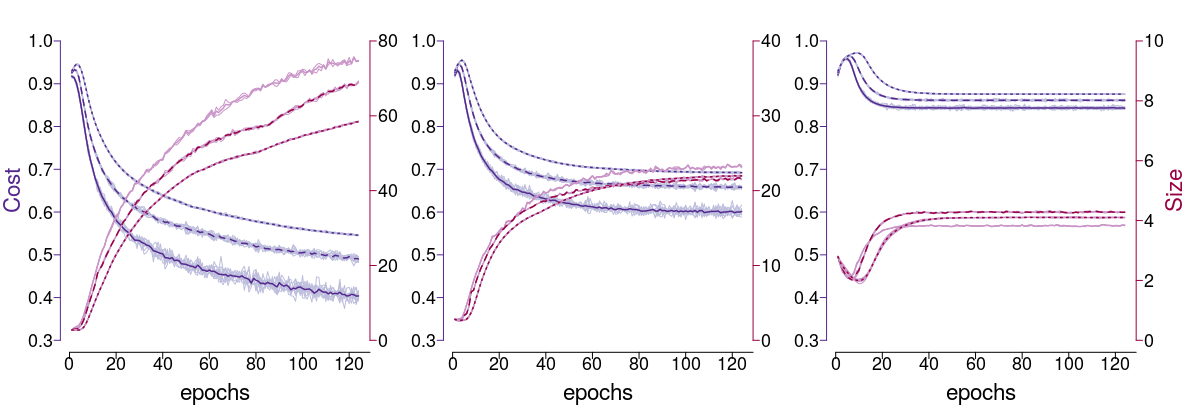}
\caption{\textbf{Embedding cost and embedding size functions.} Embedding cost and embedding size functions for the runs shown in Fig.~\ref{fig:dwt1005_ptSNE}, with perplexities $\{ 50, 200, 800 \}$ (left to right) and layers $\{ 2, 4, 8 \}$ (solid, dashed and doted lines). We depict the average embedding cost (computed from partial t-SNEs, left-hand y-axes, blue lines) and average embedding size (computed by layers, right-hand y-axes, red lines). Note the different scaling of the right-hand y-axes for size. The light-grey shadow around the solid lines show the variability of the cost and size values: the dispersion around the average cost depicts the differences in convergence among the partial t-SNEs; the dispersion around the average size depicts the differences in convergence among the layers.}
\label{fig:dwt1005_CostSize}
\end{figure}

As the mapping is improved, the embedding cost (Eq.~\ref{eq:normalized_cost}) decreases, and the embedding size (computed as the length of the diagonal of the embedding space) increases to better accommodate moderate and large dissimilarities. The depiction of the cost and the size of the embedding at the end of each epoch (Fig.~\ref{fig:dwt1005_CostSize}) allows us to assess the stability of the output.

\begin{figure}[!t]\centering
\includegraphics[width=10cm, height=7.5cm]{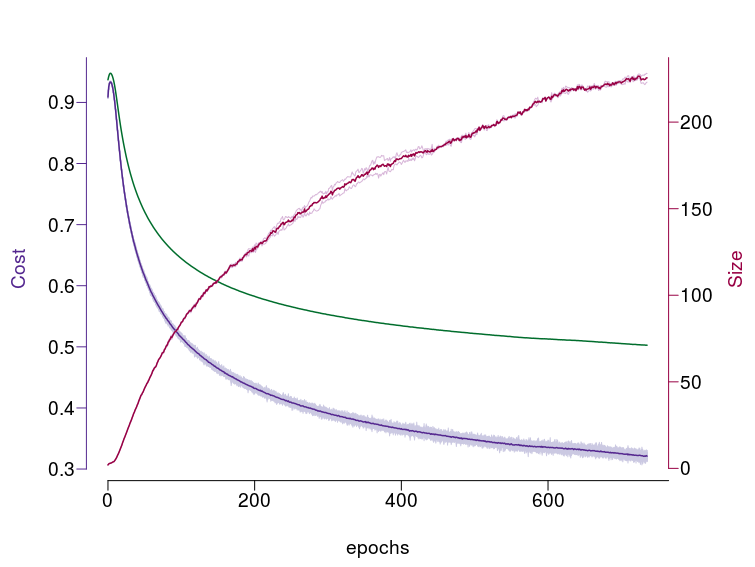}
\caption{\textbf{Comparison of the \textit{average} and \textit{global} embedding costs.} \textit{Average} embedding cost (computed from partial t-SNEs, blue line) and \textit{global} embedding cost (computed for the whole data set, green line). We also show the \textit{average} embedding size (red line). These functions correspond to the output of ptSNE for the MNIST data set (shown in Fig. \ref{fig:rounds}). }
\label{fig:globalCost}
\end{figure}

Evaluating the cost function (Eq.~\ref{eq:normalized_cost}) is an expensive operation implying the computation of the joint distributions $P$ and $Q$ for the whole data set ($\boldsymbol{O}\left(n^2\right)$). We take advantage of our parallelized implementation computing the cost function as an \textit{average} of the embedding cost of the partial t-SNEs, ($\boldsymbol{O}\left(z^2\right)$, $z \ll n$), which can be computed independently. Being this computation much faster, the \textit{average} cost is qualitatively equivalent to the \textit{global} (\textit{i.e.} considering the full data set) embedding cost (Fig. \ref{fig:globalCost}). Quantitatively, we observe a significant difference. This difference is an effect of the \textit{big crowding problem}. As $n \gg z$ the distribution of similarities for the whole data set is more homogeneous on average and is more difficult to be fairly represented in 2D. Thus, the larger is the data set size, the lower it is the power of the t-SNE to fairly represent the similarities and the cost value tends to be higher. Herein, our pseudo-normalized cost function \ref{eq:cost_function} will never equalize this effect.

Figure~\ref{fig:dwt1005_CostSize} shows the embedding cost and size functions for different parametric configurations. We observe the following:

\begin{itemize}

\item Effects on the embedding cost: while higher perplexities and higher number of layers improve the embedding (Fig.~\ref{fig:dwt1005_ptSNE}), the cost increases with both (Fig.~\ref{fig:dwt1005_CostSize}). The reason is that increasing either of them, perplexity or layers, leads to a more uniform distribution of similarities: (i) higher perplexities imply local kernels with higher variances $\sigma_{i}$, (ii) more layers entail more data points in each partial t-SNE. Therefore, although the embedding is better, the low-dimensional distribution must be closer to a uniform, hence, the embedding cost is larger.

\item Effects on the embedding size: the size of the embedding area decreases with increasing values of perplexity (Fig.~\ref{fig:dwt1005_CostSize}). Again, the reason is that higher perplexities result in more homogeneous similarities and, due to normalization, they must be lower on average, hence the embedding space is smaller and its stationary size is reached earlier. The effect of the layers on the size of the embedding is not so clear. This is due to the inter-epochs mixing phase: sooner or later, all data points end up playing its roll in all partial t-SNEs, independently of the thread-size.

\item Effects on the variability of cost and size: the less the number of layers, the less it is the overlapping among the threads (the amount of shared information) and, consequently, the more it is the dispersion (light-grey shadow around the solid lines) of both, the cost and the size of the embedding. Also, low perplexities imply that high pair-wise distances are not so well determined. As similarities are recomputed at each epoch and at each partial t-SNE, for a different chunk of data, the positioning of the most extreme data points is more unstable.

\end{itemize}

\begin{figure}[tp]\centering
\begin{tabular}{c}
\includegraphics[width=10.0cm, height=7.5cm]{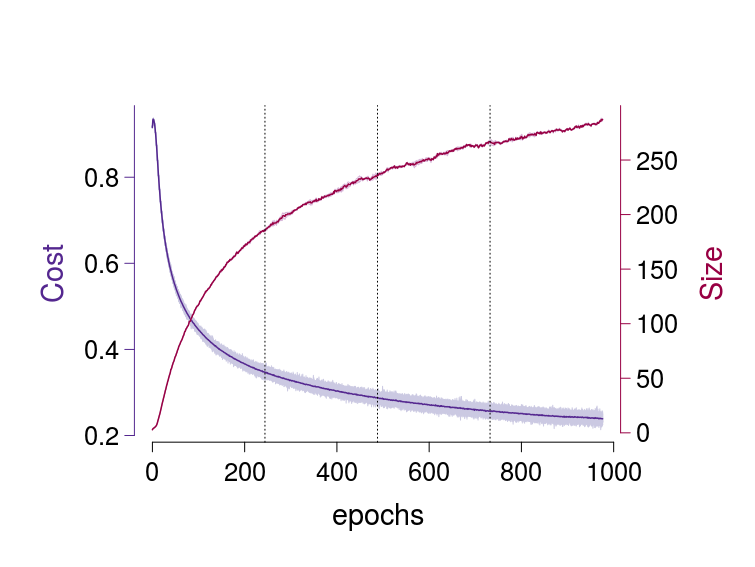}
\\
\includegraphics[width=12.0cm, height=10.2cm]{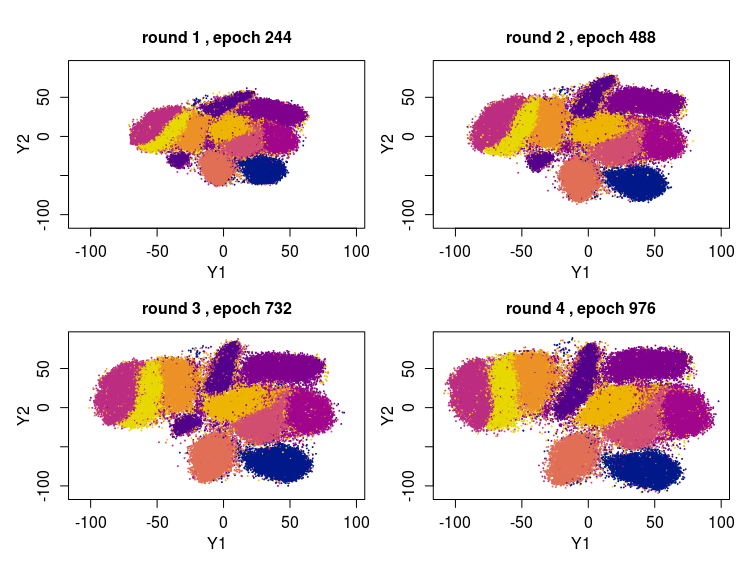}
\end{tabular}
\caption{\textbf{ptSNE rounds} ptSNE output corresponding to the MNIST (handwritten digits) data set with $n=60000$ observations and 784 dimensions. Raw data is whitened and we use the first 30 dimensions as input data. ptSNE runs with 120 threads, 2 layers ($z=1000$), 4 rounds (244 epochs per round, 32 iterations per epoch) and $ppx=400$. We compare the results after each round. Top: Embedding cost (blue lines) and embedding size (red lines) as a function of the epochs. Dotted lines indicate the end of each round. Bottom 4 panel: State of the embedding after each round. Colours depict class labels.}
\label{fig:rounds}
\end{figure}

We can further assess the robustness and stability of our solution by increasing the number of rounds (Fig.~\ref{fig:rounds}). We show the output of the ptSNE for the MNIST data set ($n=60000$) with $threads=120,\, layers=2,\, rounds=4,\, ppx=400$ and 244 epochs per round. We plot the embedding cost and size functions (top panel) indicating the rounds with dotted lines. The 4 panels at the bottom show the state of the embedding at the end of each round (colours depict class labels). After the first round (Fig.~\ref{fig:rounds}, $round=1,\, epoch=244$), the main shape of the mapping is almost defined except for one class (split into two clusters). After the fourth round, this class becomes a unified single cluster (Fig.~\ref{fig:rounds}, $round=4,\, epoch=976$). The cost and size functions end up almost stable though still slightly improving. This improvement responds uniquely to the fact that the algorithm achieves a better matching of large dissimilarities by enlarging the low dimensional embedding. The stability of the cost and size functions is a clear sign that the mapping is stable and its shape will hardly change any more.

\subsection*{ptSNE computation times}

\begin{figure}[tp]\centering
\begin{tabular}{c}
\includegraphics[width=16cm, height=12.5cm]{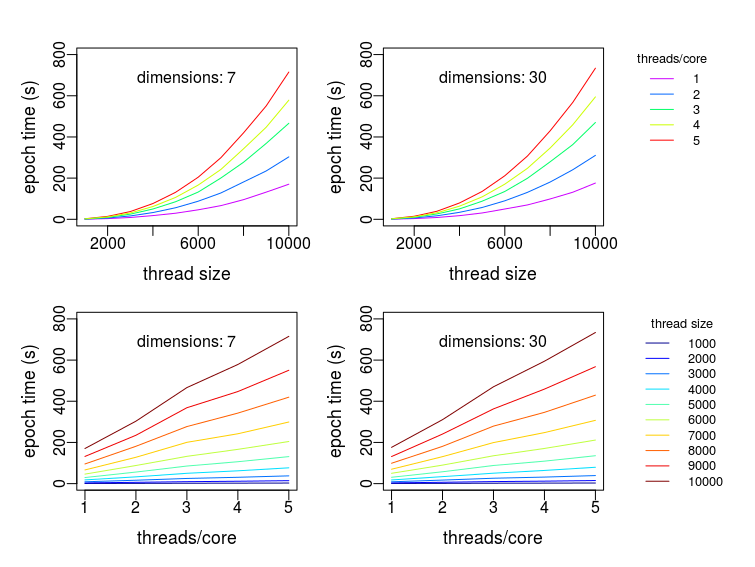}
\\
\includegraphics[width=9.0cm, height=6.5cm]{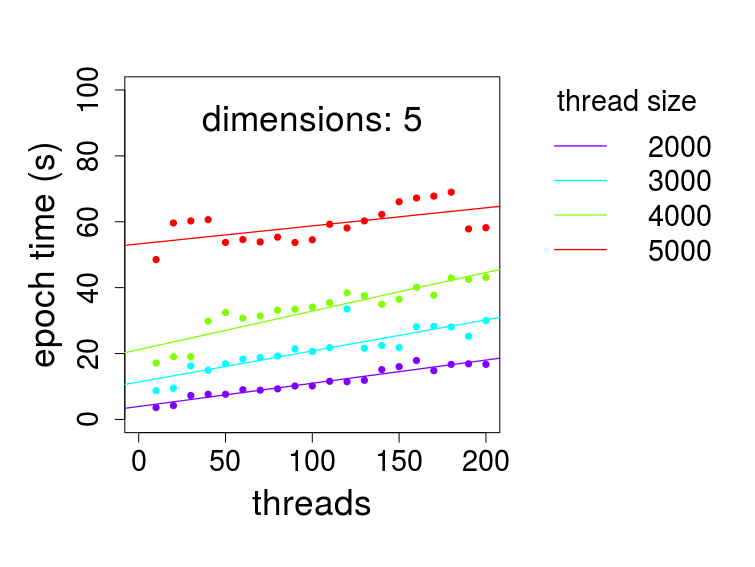}
\end{tabular}
\caption{\textbf{ptSNE computation times.} Computation times given in seconds per epoch. Top 4-panel: \textit{SOCK} (intra-node) parallelization, two data sets with 7 input dimensions (left) and 30 input dimensions (right). Computation times given as a function of thread-size $z$ and multi-threading ratio $threads/core$. Bottom panel: \textit{MPI} (inter-node) parallelization (data set with 5 input dimensions). Computation times given as a function of thread-size $z$ (by colors) and number of threads.}
\label{fig:computationTimes}
\end{figure}

Current on-the-shelf t-SNE in \proglang{R} \citep{Krijthe:2015}, python \citep{sklearn:2011} and Matlab \citep{Maaten:2009a} include accelerated versions of the algorithm based on ball-tree approaches: (i) a sparse approximation of the similarities between the input objects using vantage-point trees \citep{Yianilos:1993}, and (ii) a Barnes-Hut approximation \citep{Barnes:1986} to compute the gradient function. These are powerful approximations that, at the cost of misrepresenting the similarity distribution in both, the high and low dimensional spaces, dramatically reduce the time and space complexity of the algorithm to $\boldsymbol{O}\left(n\,log n\right)$ computation and $\boldsymbol{O}\left(n \right)$ memory. ptSNE implements the exact computation of the gradient function. The exact computation is of $\boldsymbol{O}\left(n^2\right)$ but, thanks to parallelization, is reduced to $\boldsymbol{O}\left(z^2\,\sqrt{n}\right)$ computation time and $O\left(z^2\right)$ memory space, where $z$ is the thread-size, $z\ll n$, and $\sqrt{n}$ stands for the number of epochs per round.

We have used \proglang{C++} (\pkg{Rcpp}, \pkg{RcppArmadillo} packages \citep{Rcpp:2013, RcppArmadillo:2014}) and shared memory (\pkg{bigmemory} package \citep{bigmemory:2013}) to improve the memory needs and the computation times of the most expensive parts of the protocol. Parallelization is implemented at low level by means of the \pkg{snow} package \citep{snow:2016} allowing both \textit{SOCK} (intra-node) parallelization and also \textit{MPI} (inter-node) parallelization. Nonetheless, ptSNE is still far from the computational efficiency of approximated t-SNE: it runs comfortably even with standard memory resources (16 to 64GB), although at the cost of considerable times for large data sets. Therefore, the benefits of ptSNE must not be considered in terms of computational time but in terms of robustness and exactness of the solution. In Fig.\ref{fig:computationTimes} we show the computation times for a range of parametric setups using three different data sets (GMM5, GMM7 and MNIST) with 5, 7 and 30 input dimensions respectively. The computation times are given in seconds per epoch (\textit{i.e.} the expected overall time is the epoch running time times $\sqrt{n}$ epochs per round).

Using \textit{SOCK} parallelization (Fig.~\ref{fig:computationTimes}, top 4-panel), the computation times increase quadratically with the thread-size ($z=n\,*\,layers/threads$) but almost linearly with the multi-threading ratio ($threads/core$), while the number of input dimensions has almost no effect. As an example, given a data set with $n=300000$ and hardware resources limited to 60 cores: (i) using 60 threads (\textit{i.e.} 1 $thread/core$) and 2 layers (\textit{i.e.} $z=10000$) takes about 170 seconds/epoch (Fig.~\ref{fig:computationTimes}, top 4-panel, purple line); (ii) using 180 threads (\textit{i.e.} 3 $threads/core$) and 3 layers (\textit{i.e.} $z=5000$) takes about 85 seconds/epoch (Fig.~\ref{fig:computationTimes}, top 4-panel, green line). As one single round takes 547 epochs ($epochs=\sqrt{n}$) the first strategy yields around 26 hours of computation per round while the second strategy yields only 13 hours/round.

Using \textit{MPI} parallelization (Fig.~\ref{fig:computationTimes}, bottom panel), our measurements of computation times are not so consistent as they include the message passing times between master and worker processes which are affected by the overall load of the system. However, the computation times increase roughly linearly with the number of threads (\textit{i.e.} the number of physical cores here, as we did not use multi-threading in this example). In other words, using \textit{MPI} with no multi-threading, and a given parametric configuration, each additional physical core allows increasing the data set size by one thread-size, and this results in a linear increase in computation time. Following with the previous example with $n=300000$, using 200 cores and 2 layers (\textit{i.e.} $z=3000$) takes about 30 seconds per epoch (Fig.~\ref{fig:computationTimes}, bottom panel, light-blue line), that is around 4.5 hours of computation per round.

\subsection*{Clustering}

\begin{figure}[!p]\centering
\begin{tabular}{cc}
\includegraphics[width=7cm, height=6cm]{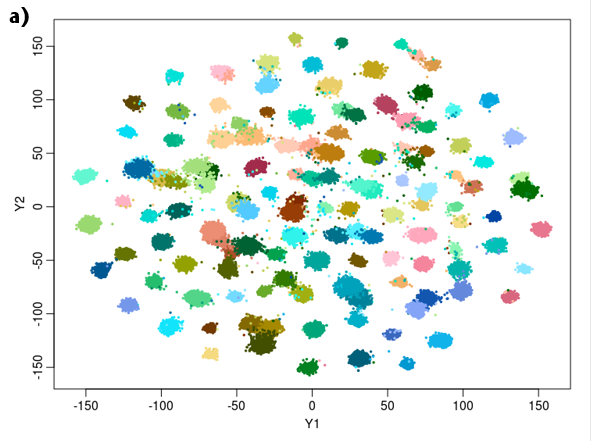}
&
\includegraphics[width=7cm, height=6cm]{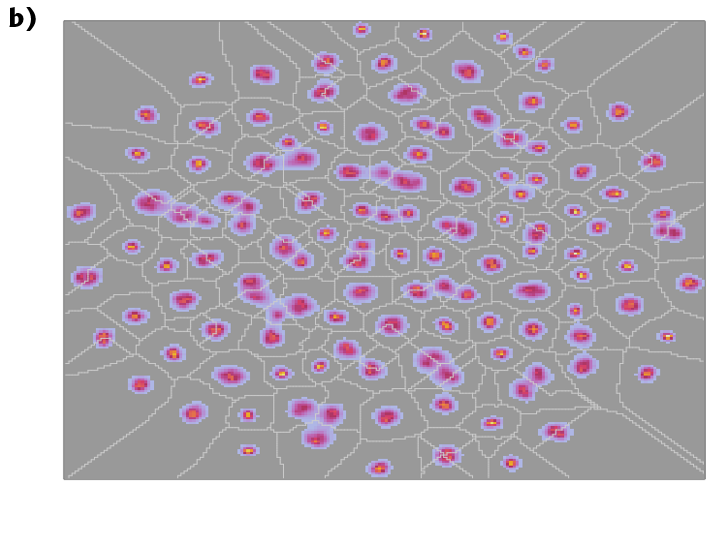}
\\
\includegraphics[width=7cm, height=6cm]{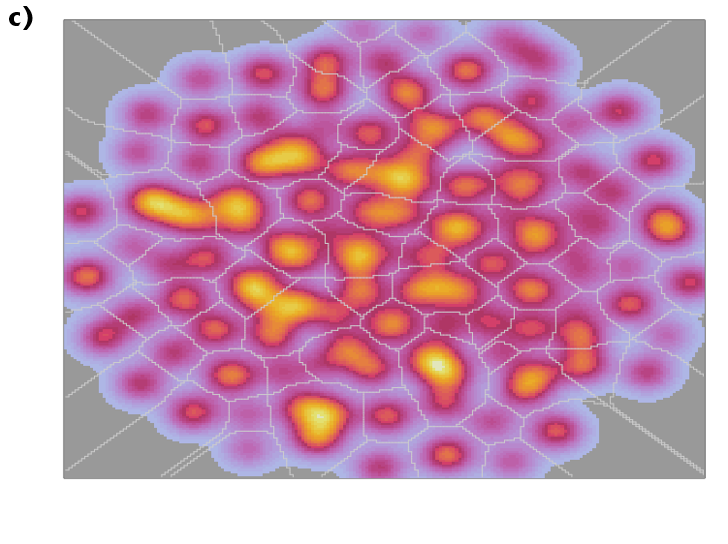}
&
\includegraphics[width=7cm, height=6cm]{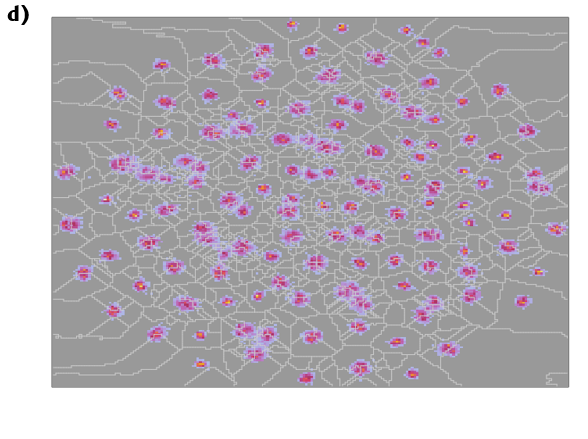}
\end{tabular}
\caption{\textbf{Fix-bandwidth \textit{vs} perplexity-adaptive KDE.} Clustering results for the GMM7 data set. This is a synthetic data set with $n=63998$ observations sampled from a 7 dimensional GMM with 128 Gaussian components. a) ptSNE output ($perplexity=100, \,threads=64, \,layers=2$), colours show the original Gaussian components, illustrating the correctness of the embedding. b) paKDE output ($perplexity=100$). Bottom: Density estimation using $kde2d\left(\right)$ (\pkg{MASS} \proglang{R}-package, \citet{MASS:2002}) with fix bandwidth c) $h=\{ 34.35,\,34.95 \}$ (rule-of-thumb value) and d) $\{ 1.7, 1.7 \}$ (paKDE mean bandwidth value, Table~\ref{tbl:sigma_summary}). In b), c) and d) the white lines depict the boundaries of the clusters found by the WTT algorithm: b) 166, c) 85, and d) 585 clusters out of 128 original components.}
\label{fig:GMM7}
\end{figure}

We use the GMM7 data set to compare the clustering that results from paKDE and a fixed bandwidth kernel density estimation ($kde2d\left(\right)$) included in the \pkg{MASS} \proglang{R}-package \citep{MASS:2002}. For the later, we test two different bandwidth values: (i) $h=\{ 34.35,\,34.95 \}$, a rule-of-thumb bandwidth suggested as default value \citep{MASS:2002}, and (ii) $h=\{ 1.7, 1.7 \}$, the mean local bandwidth computed by paKDE (summarized in Table~\ref{tbl:sigma_summary}). For ptSNE and paKDE we used $ppx=100$ (\textit{i.e.} equivalent definitions of similarity in the high and low dimensional spaces). In Fig.~\ref{fig:GMM7} we show the output of ptSNE (a)) and the clusterings that result from: paKDE+WTT (b)), $kde2d(h=\{ 34.35,\,34.95 \})$+WTT (c)), and $kde2d(h=\{ 1.7, 1.7 \})$+WTT (d)).

\begin{table}\center
\begin{tabular}{cccccc}
Min. & 1st Qu. & Median & Mean & 3rd Qu. & Max. \\
\hline
0.938 & 1.363 & 1.574 & 1.703 & 1.928 & 6.223 \\
\hline
\end{tabular}
\caption{Summary of paKDE bandwidths computed for the GMM7 data set with $perplexity=100$.}
\label{tbl:sigma_summary}
\end{table}

The GMM7 presents a significant degree of overlap for some of the 128 Gaussian components. Consequently, ptSNE is not able to separate all of them. That being said, a fix bandwidth density estimation with $h\approx 34$ (Fig.~\ref{fig:GMM7}c) is excessively smoothed and accumulates too much density at the centre part of the embedded area. The clustering yields only 85 clusters missing many of the original Gaussian components. Using a lower fix bandwidth $h=1.70$ (Fig.~\ref{fig:GMM7}d), KDE forms needle peaks for every small aggregation of mapped data points and the clustering yields 585 clusters, most of which are irrelevant. The paKDE output (Fig.~\ref{fig:GMM7}b) is adequately tightened to the ptSNE embedding and does not show any bias towards the centre of the embedding area. Such density landscape favours the detection of low-density clusters or clusters with narrow aisles in between. As a result, the clustering yields 166 clusters, still splitting some of the 128 original components, but much closer to the underlying GMM. These properties benefit a potential quantitative interpretation of the clustering densities.

\begin{figure}[!p]\centering
\begin{tabular}{c}
\end{tabular}
\begin{tabular}{cc}
\includegraphics[width=7.0cm, height=5.8cm]{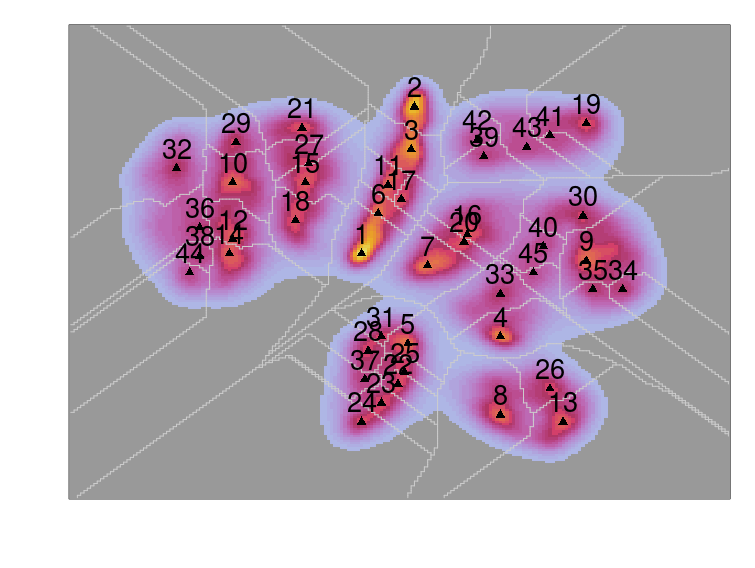}
&
\includegraphics[width=7.0cm, height=5.8cm]{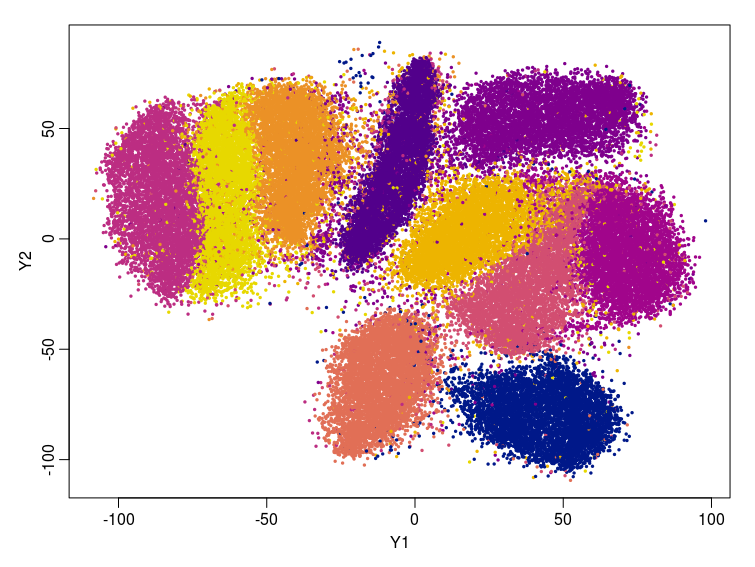}
\\
\includegraphics[width=7.0cm, height=1.1cm]{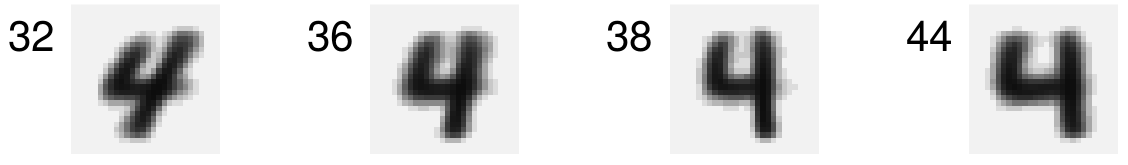}
&
\\
\includegraphics[width=7.0cm, height=1.1cm]{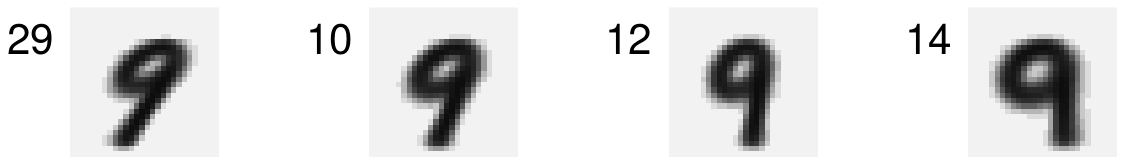}
&
\\
\includegraphics[width=7.0cm, height=1.1cm]{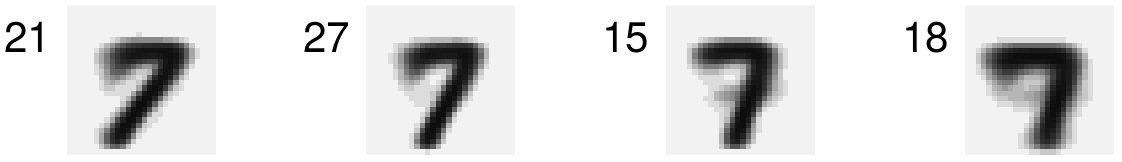}
&
\\
\includegraphics[width=7.0cm, height=1.1cm]{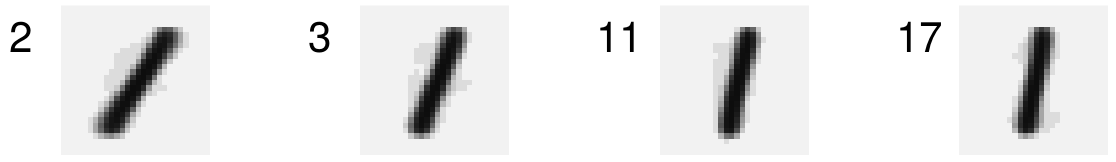}
&
\includegraphics[width=7.0cm, height=1.1cm]{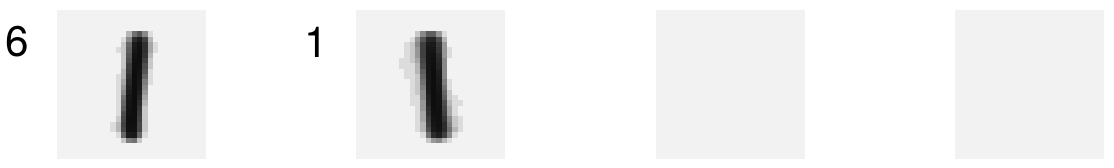}
\\
\includegraphics[width=7.0cm, height=1.1cm]{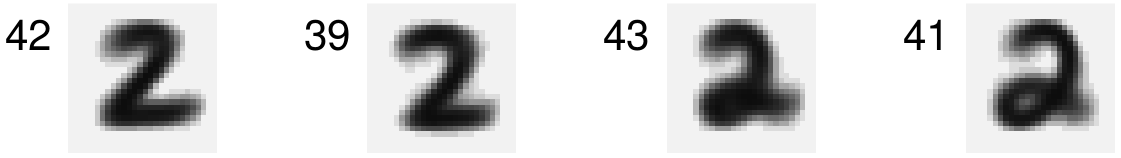}
&
\includegraphics[width=7.0cm, height=1.1cm]{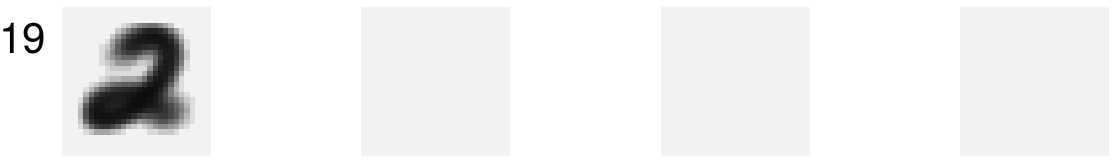}
\\
\includegraphics[width=7.0cm, height=1.1cm]{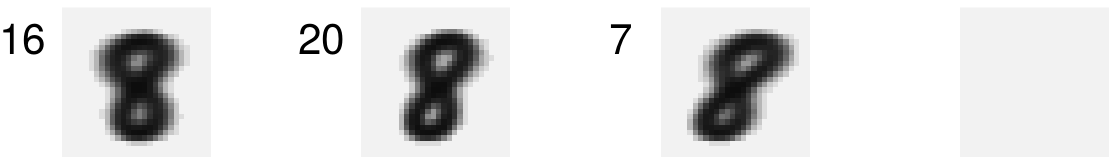}
&
\\
\includegraphics[width=7.0cm, height=1.1cm]{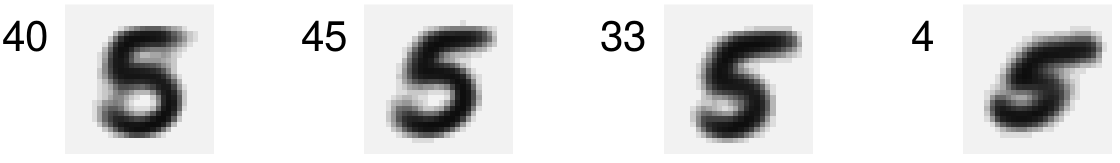}
&
\\
\includegraphics[width=7.0cm, height=1.1cm]{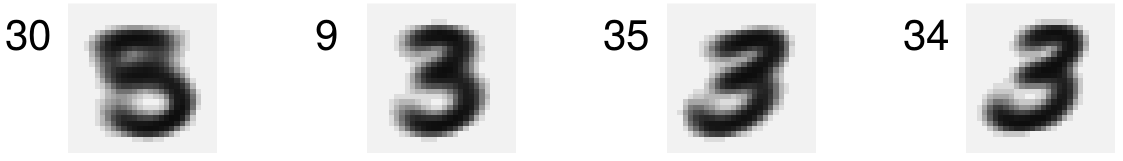}
&
\\
\includegraphics[width=7.0cm, height=1.1cm]{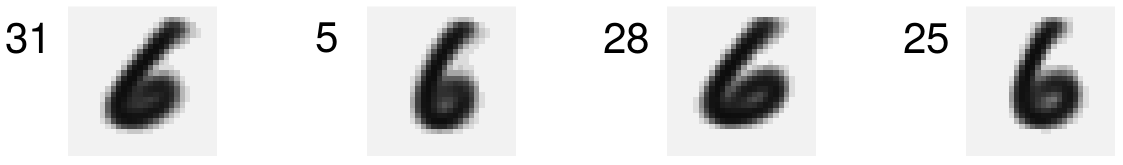}
&
\includegraphics[width=7.0cm, height=1.1cm]{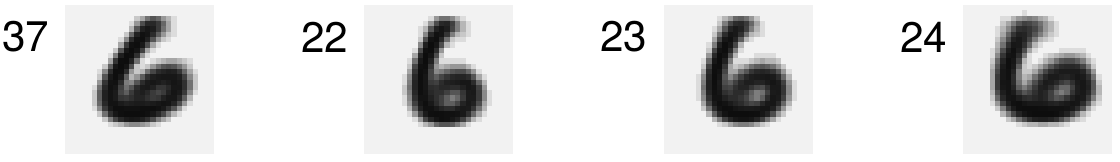}
\\
\includegraphics[width=7.0cm, height=1.1cm]{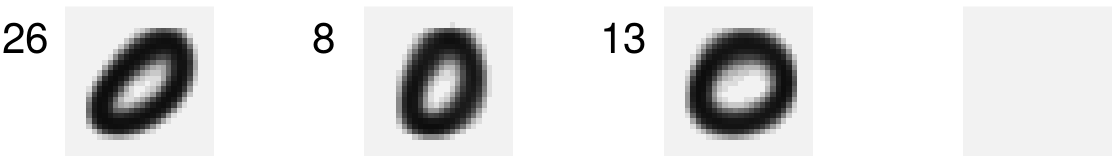}
&
\end{tabular}
\caption{\textbf{ptSNE+paKDE+WTT} MNIST data set clustering result. ptSNE performed as described in Fig.~\ref{fig:rounds}. We use $perplexity=400$ for both ptSNE and paKDE. Top left: paKDE+WTT clustering (45 clusters). Top right: ptSNE output, colours depict class labels. Bottom rows: Average grey intensity per pixel of the set of images in each cluster (numbers in the top-left corner of the images indicate number of cluster as shown in the top-left plot).}
\label{fig:MNIST}
\end{figure}

We also show the clustering of the MNIST data set using ptSNE+paKDE+WTT (Fig.~\ref{fig:MNIST}). The ptSNE is parameterized as in Fig.~\ref{fig:rounds}, with $perplexity=400$ in both ptSNE and paKDE. The result is a fine-grain hierarchical clustering (45 clusters) showing an excellent trade-off between global and local structure: classes are well separated and several subclasses are identified in each, and the overall arrangement of classes and subclasses is seamlessly consistent throughout the embedding (Fig.~\ref{fig:MNIST}, bottom rows).

(More examples at \url{http://gitlab.ceab.csic.es/jgarriga/bigmap_collection}.)

\section*{Conclusions}

We present a \textit{mapping method} (MM) particularly suited for large-scale structured data (LSSD). The MM is a multi-step protocol to perform an unsupervised clustering over a low dimensional representation of the data. The clusters are found by estimating a density function over the low dimensional embedding of the data and afterwards segmenting the density landscape following the river beds (water-tracks) in it.

The dimensionality reduction of the data is based on the ptSNE algorithm, a parallelized implementation of the well-known t-Stochastic Neighbouring Embedding (t-SNE) \citep{Maaten:2008} that improves the suitability of the t-SNE for LSSD. Our approach is not grounded on high-performance hardware developments (\textit{e.g.} t-SNE-CUDA, \cite{Chan:2018}, \cite{cudatsne:2018}) but on a reconsideration of the algorithm itself from a big data perspective. Based on the assumption that massive amounts of data convey large amounts of redundant evidence, we breakdown the t-SNE into partial t-SNEs that we adequately parameterize and combine to obtain a convergent global solution. This parallelization scheme adds some extra computation effort related to task organization by a master and a set of worker processes. Nonetheless, we expand the computational limits of the algorithm alleviating the main drawback of non-parallelized t-SNE exact implementations when dealing with LSSD: the algorithmic complexity of the ptSNE is determined by the thread-size $z$ of the partial t-SNEs, significantly smaller than the data set size $n$. The time complexity of the ptSNE is quadratically decreased from $n^2$ to $z^2\,\sqrt{n}$ ($z \ll n$). Likewise, the space complexity is strongly alleviated because there is no need to compute a complete distance matrix: only partial distance matrices are computed at each thread. Additionally, we minimize the \textit{big crowding problem} (the increasing difficulty to fairly represent pairwise similarities for LSSD of increasing size when using a finite mass probability distribution). Also, as a result of starting with multiple random initial positions, the final solution is not so much dependent on the starting conditions. In summary, although the computational efficiency of ptSNE is still far from current on-the-shelf accelerated implementations of t-SNE, its benefits are clear in terms of robustness and exactness of the solution.

Our results show that ptSNE converges to a global and stable solution as long as: (i) the thread-size is large enough so that each chunk of data conveys sufficient information about the global structure in the data; (ii) the epoch length is not too large (low number of iterations per epoch) to avoid too divergent solutions at each thread; and (iii) the number of epochs is large enough to reach a stable solution. The default settings that we describe are broadly conservative to fulfill these conditions.

We estimate the density function over the low-dimensional embedding using the perplexity-adaptive Kernel Density Estimation (paKDE) algorithm. The novelty in this adaptive KDE is to use a perplexity based approach to find the local bandwidths. This approach allows us to get accurate density estimations out of the embedded data while adequately couples paKDE with ptSNE, linking both steps by a single backbone idea: the pairwise similarities, in the high dimensional space (ptSNE), and in the low dimensional space (paKDE). Because of this backbone link, we can think globally about similarity as a single value of perplexity (understanding that we may be willing to use different values of perplexity for paKDE to get a finer/coarser clustering).

For the segmentation of the density landscape, we introduce the water-track transform (WTT) algorithm, a variant of the rain-falling watershed transform that identifies clusters as peaks, and their influence areas, in the density landscape by following the river beds in the valleys.

We wrapped our mapping protocol in the \pkg{bigMap} \proglang{R} package. The package includes a set of high-level functions to easily perform the whole protocol and is complemented with some visualization and analysis tools to ease the qualitative and quantitative assessment of the clustering. The package is mainly intended to work remotely, launching batch processes on high-performance computing platforms, but allows also working interactively, within the R's environment using moderate hardware.

Our main concern has been on the convergence and robustness of the implementation through the whole mapping protocol while trying to achieve a reasonable efficiency. Where possible, parallelization is implemented by means of the \textit{snow} package \citep{snow:2016} allowing both \textit{SOCK} (intra-node) parallelization and \textit{MPI} (inter-node) parallelization. The most expensive computation parts are coded in \proglang{C++} using the \proglang{R} interfaces \pkg{Rcpp} and \pkg{RcppArmadillo} \citep{Rcpp:2011, Rcpp:2013}. Memory resources are managed by means of the \pkg{bigmemory} package \citep{bigmemory:2013}. Based on the Boost Interprocess \proglang{C++} library, the \pkg{bigmemory} package provides platform-independent support for massive matrices that may be shared across R processes, using shared memory and memory-mapped file. This set up provides substantial speed and memory efficiencies while maintaining the flexibility and power of \proglang{R}'s rich statistical programming environment.

There is plenty of room to improve the efficiency of our implementation. The computation time of ptSNE would benefit from approximations like Barnes-Hut \citep{Maaten:2014} or Fast Fourier Transform-accelerated Interpolation-based t-SNE \citep{Linderman:2017}, but the consequences of implementing these approximations must be carefully analysed in terms of potential convergence issues. Improvements can also come from the combination of algorithmic and hardware optimization strategies, for example, linking our parallelized scheme with powerful CUDA implementations as in \cite{Chan:2018, cudatsne:2018}. A further improvement is to integrate our parallelized approach into a real-time tool for progressive visual analytics \citep{Mulbacher:2014, Stolper:2014} following the line recently described in \cite{Maaten:2017}, a user steerable implementation of the t-SNE algorithm.

\section*{Computational details}

The results presented in this paper were obtained using \proglang{R}~3.5.1 and \pkg{bigMap}~1.9.7, running on the high-performance computing cluster at the Computational Biology Lab (CEAB-CSIC) \url{http://www.ceab.csic.es/en/services/computational-biology-lab/} (Table~\ref{tbl:CBLab}). \proglang{R} itself and all packages used are available from the Comprehensive \proglang{R} Archive Network (CRAN) at \url{https://CRAN.R-project.org/}.

\begin{table}[!t]
\small
\begin{tabular}{ccccccc}
nodes & model & CPU & cores & fr.(MHz) & RAM & OS (bits)\\
\hline
5 & PowerEdge R420 & Intel(R) Xeon(R) E5-2450L & 16 & 1800 & 161G & 64 \\
7 & PowerEdge R430 & Intel(R) Xeon(R) E5-2650 & 20 & 2300 & 193G & 64 \\
1 & PowerEdge R815 & AMD Opteron(tm) 6380 & 64 & 2500 & 515G & 64 \\
\hline
\end{tabular}
\caption{\textbf{High-performance computing cluster at the Computational Biology Lab (CEAB-CSIC)}. Technical specifications.}
\label{tbl:CBLab}
\end{table}

\section*{Acknowledgments}

We acknowledge members from the Theoretical and Computational Ecology lab to provide comments on previous versions of the MS, and insights as users of beta versions of the \pkg{bigMap} \proglang{R}-package. This work was supported by the Spanish Ministry (MINECO, Grant CGL2016-78156-R), and the Max Planck Institute for Ornithology (MPIO, Germany). The high-performance computation cluster at the Computational Biology Lab (CEAB-CSIC) was supported by the Spanish Ministry (MINECO, CSIC13-4E-1999).

\bibliography{bigMap}

\end{document}